\documentclass{article}

\PassOptionsToPackage{numbers, compress}{natbib}

\usepackage[preprint]{neurips_2025}

\usepackage[utf8]{inputenc} 
\usepackage[T1]{fontenc}    
\usepackage{hyperref}       
\usepackage{url}            
\usepackage{booktabs}       
\usepackage{amsfonts}       
\usepackage{nicefrac}       
\usepackage{microtype}      
\usepackage{xcolor}         

\usepackage[export]{adjustbox}

\usepackage{amssymb}
\usepackage{amsmath}
\usepackage{subcaption}
\usepackage{enumitem}
\usepackage{tcolorbox}
\usepackage[capitalize,noabbrev]{cleveref}
\usepackage{CJKutf8}

\newcommand{\vx}{\boldsymbol{x}}

\newcommand{\vtheta}{\boldsymbol{\theta}}
\newcommand{\vp}{\boldsymbol{p}}
\newcommand{\vr}{\boldsymbol{r}}
\newcommand{\vv}{\boldsymbol{v}}

\title{LLaDA-V: Large Language Diffusion Models with \\ Visual Instruction Tuning}

\author{%
  Zebin You$^{1, 2, 3}$\thanks{Work done during an internship at Ant Group.}, Shen Nie$^{1, 2, 3}$, Xiaolu Zhang$^4$, Jun Hu$^{4}$, Jun Zhou$^4$, \\ 
  \textbf{Zhiwu Lu}$^{1, 2, 3}$, \textbf{Ji-Rong Wen}$^{1, 2, 3}$, \textbf{Chongxuan Li}$^{1,2,3}$\thanks{Correspondence to Chongxuan Li.}\\
  $^1$ Gaoling School of AI, Renmin University of China
  $^2$ Beijing Key Laboratory of Research on \\ Large Models and Intelligent Governance 
  $^3$ Engineering Research Center of  Next-Generation \\ Intelligent Search and Recommendation, MOE 
  $^4$ Ant Group \\ 
}

\begin{document}

\maketitle

\begin{abstract}
In this work, we introduce \textbf{LLaDA-V}, a purely diffusion-based Multimodal Large Language Model (MLLM) that integrates visual instruction tuning with masked diffusion models, representing a departure from the autoregressive paradigms dominant in current multimodal approaches. Built upon LLaDA, a representative large language diffusion model, LLaDA-V incorporates a vision encoder and MLP connector that projects visual features into the language embedding space, enabling effective multimodal alignment. Our empirical investigation reveals several intriguing results: First, LLaDA-V demonstrates promising multimodal performance despite its language model being weaker on purely textual tasks than counterparts like LLaMA3-8B and Qwen2-7B. When trained on the same instruction data, LLaDA-V is highly competitive to LLaMA3-V across multimodal tasks with better data scalability. It also narrows the performance gap to Qwen2-VL, suggesting the effectiveness of its architecture for multimodal tasks. Second, LLaDA-V achieves state-of-the-art performance in multimodal understanding compared to existing hybrid autoregressive-diffusion and purely diffusion-based MLLMs. Our findings suggest that large language diffusion models show promise in multimodal contexts and warrant further investigation in future research. Project page and codes: \url{https://ml-gsai.github.io/LLaDA-V-demo/}.
\end{abstract}

\section{Introduction}

Multimodal Large Language Models (MLLMs) are capable of processing multiple input modalities—including images~\cite{liu2023visual, liu2024improved, li2024llava, chen2024internvl, wang2024qwen2, team2024chameleon}, audio~\cite{ding2025kimi, chu2023qwen, ghosh2024gama}, and video~\cite{wang2025internvideo2, chen2024sharegpt4video, zhang2024video}—alongside text, and can generate natural language responses that follow given diverse instructions. Despite significant advancements in MLLMs, most existing approaches predominantly rely on autoregressive models~\cite{radford2018improving,radford2019language, brown2020language, touvron2023llama, touvron2023llama2, grattafiori2024llama,yang2024qwen2,li2023textbooks,bi2024deepseek}, leaving substantial room for exploring alternative probabilistic modeling approaches. 

Recent attempts to incorporate diffusion models~\cite{sohl2015deep,ho2020denoising,song2020score,hoogeboom2021argmax,austin2023structured} into MLLMs have predominantly adopted one of two strategies: either leveraging autoregressive models to provide strong language modeling capabilities~\cite{bao2023one,xie2024show,zhou2024transfusion,ma2024janusflow,tong2024metamorph,kou2024orthus}, or employing discrete diffusion-based approaches with limited language modeling capacity, which consequently leads to suboptimal performance~\cite{swerdlow2025unified,li2024dual}. 

Encouragingly, recent advances in discrete diffusion models~\cite{hoogeboom2021argmax,austin2023structured,campbell2022continuous,he2022diffusionbert,Sun2022ScorebasedCD,loudiscrete,shi2024simplified,sahoo2024simple,ou2024your,nie2025large,you2025effective} have shown promising potential to overcome these limitations. In particular, LLaDA~\cite{nie2025large} has demonstrated performance competitive with LLaMA3-8B-Instruct~\cite{grattafiori2024llama} through large-scale pre-training and SFT, while retaining favorable scaling properties. Nevertheless, while LLaDA has shown remarkable progress in language modeling, its capabilities and potential in multimodal understanding remain largely unexplored. Therefore, this naturally raises a key research question: Can an purely diffuision based MLLM (both training and sampling) achieve performance compared to autoregressive-based models? 

In this paper, we explore how to effectively extend large language diffusion models to encompass strong multimodal understanding capabilities, focusing on the visual instruction tuning framework~\cite{liu2023visual}, which has demonstrated remarkable effectiveness across various autoregressive-based MLLMs. In particular, we introduce a vision encoder (e.g., SigLIP 2~\cite{tschannen2025siglip}) and an MLP connector to map visual features into the LLaDA language embedding space, allowing joint processing of visual and textual inputs. Furthermore, we extend LLaDA's training objective to handle multi-turn multimodal dialogues, investigate various attention mechanism structures, adapt inference procedures for multimodal conversations, and develop a multi-stage training strategy. These comprehensive investigations result in \textbf{LLaDA-V}, a purely diffusion-based MLLM. 

\begin{figure}[t!]
    \centering
    \begin{subfigure}[b]{0.49\textwidth}
        \centering
        \includegraphics[width=\textwidth]{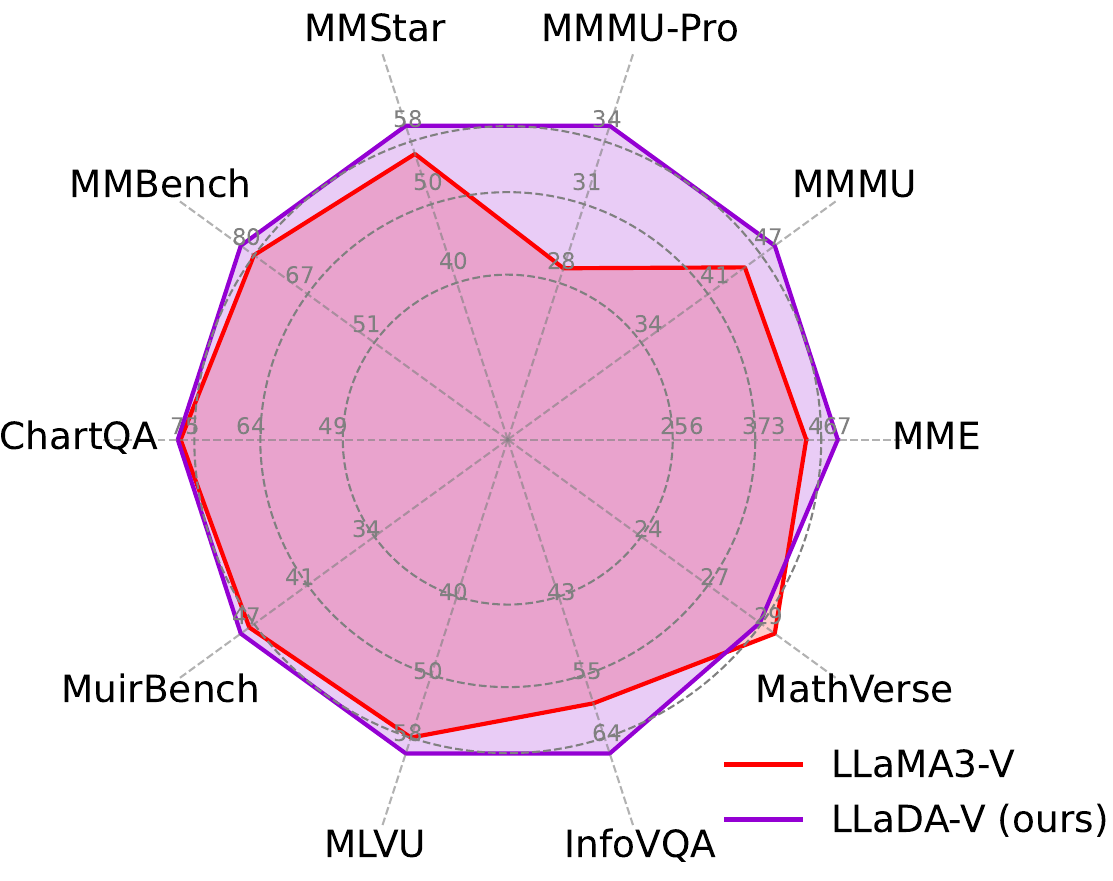}
        \caption{LLaDA-V vs. LLaMA3-V}
        \label{fig:llada_v_baseline_comparison}
      \end{subfigure}
    \hfill
    \begin{subfigure}[b]{0.49\textwidth}
        \centering
        \includegraphics[width=\textwidth]{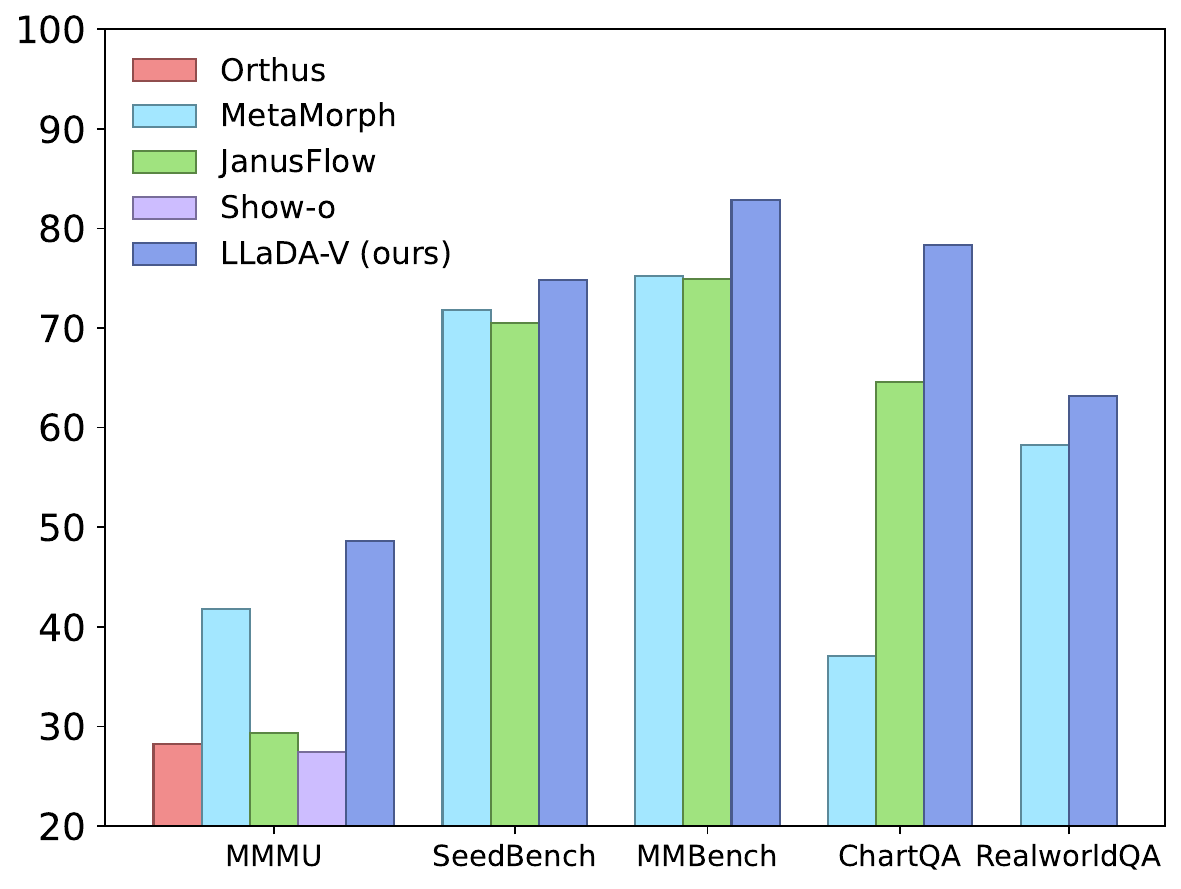}
        \caption{LLaDA-V vs. non-autoregressive MLLMs}
        \label{fig:llada_v_diffusion_comparison}
      \end{subfigure}
    \caption{\textbf{Benchmark Results.} (a) LLaDA-V demonstrates superior performance on more benchmarks compared to LLaMA3-V when trained on the same dataset, particularly excelling in multidisciplinary knowledge and mathematical reasoning tasks. (b) LLaDA-V achieves state-of-the-art performance in multimodal understanding among both hybrid autoregressive-diffusion (such as MetaMorph~\cite{tong2024metamorph} and Show-o~\cite{xie2024show}) and purely diffusion-based models.}
    \label{fig:llada_v_performance}
  \end{figure}

We first compare the data scalability of LLaDA-V to that of LLaMA3-V (our autoregressive baseline with LLaMA3-8B as the language tower) by varying the amount of instruction tuning data. LLaDA-V demonstrates stronger data scalability on several benchmarks, particularly excelling in tasks involving multidisciplinary knowledge and mathematical reasoning (see Fig.~\ref{fig:data_scaling_lladav}). 

Furthermore, we benchmark LLaDA-V against autoregressive, hybrid autoregressive-diffusion, and pure diffusion models across 18 diverse multimodal tasks. Notably, when comparing with LLaMA3-V, we observe an interesting and promising phenomenon: despite with a slightly weaker language tower, our model achieves superior performance across 11 tasks (see partial results in Fig.~\ref{fig:llada_v_performance} (a) and more details in Section~\ref{sec:exp}). Similarly, when compared to the powerful autoregressive Qwen2-VL~\cite{wang2024qwen2}, despite LLaDA being considerably weaker than Qwen2-7B, LLaDA-V narrows the performance gap significantly, achieving comparable results on some benchmarks such as MMStar~\cite{chen2024we} (60.1 vs. 60.7). Furthermore, our model achieves state-of-the-art performance compared to existing hybrid autoregressive-diffusion models~\cite{xie2024show,zhou2024transfusion,ma2024janusflow,tong2024metamorph,kou2024orthus} and pure diffusion models~\cite{swerdlow2025unified,li2024dual}(see Fig.~\ref{fig:llada_v_performance} (b)). Collectively, all these findings demonstrate not only 
 the effectiveness of the LLaDA-V framework but also the promise of diffusion models on multimodal understanding.

In summary, our key contributions are as follows:
\begin{itemize}[leftmargin=*]
    \setlength\itemindent{.3em}
\item We introduce LLaDA-V, a purely diffusion-based MLLM for multimodal understanding.
\item We demonstrate that LLaDA-V benefits from data scaling and achieves superior scalability across multiple benchmarks when compared to our autoregressive baseline, LLaMA3-V.
\item LLaDA-V achieves state-of-the-art results among both hybrid and purely diffusion-based MLLMs. 
\end{itemize}




\section{Preliminaries}
In this section, we briefly introduce large language diffusion models, which serve as the language tower in our work, and visual instruction tuning, which forms the basis of our multimodal framework.

\textbf{Large Language Diffusion Models.} Large language models (LLMs) are currently experiencing rapid development. The predominant LLMs~\cite{radford2018improving,radford2019language,brown2020language,touvron2023llama,touvron2023llama2,grattafiori2024llama,yang2024qwen2} are primarily trained using autoregressive modeling. Unlike autoregressive approaches, discrete diffusion models~\cite{sohl2015deep,hoogeboom2021argmax} offer an alternative paradigm for language modeling. Masked diffusion models~\cite{austin2023structured,campbell2022continuous}, a specific variant of discrete diffusion, have shown impressive results across multiple domains~\cite{Sun2022ScorebasedCD, loudiscrete, shi2024simplified, sahoo2024simple, ou2024your,nie2025large,nie2024scaling,campbell2024generativeflowsdiscretestatespaces,hu2024maskneed,you2025effective}.

Among them, LLaDA~\cite{nie2025large} has demonstrated comparable performance with strong AR models like LLaMA3-8B-Instruct~\cite{grattafiori2024llama}, while maintaining the unique properties of masked diffusion models. Specifically, LLaDA employs a masked diffusion process that differs fundamentally from autoregressive approaches. Formally, let $ \vx_0 = [\vx^i]_{i=1}^N $ represent a sentence comprising N tokens, and let \text{[M]} denote a special mask token. LLaDA defines a model distribution $p_{\vtheta}(\vx_0)$ through a forward and a reverse process. In the forward process, LLaDA first samples a time step $t$ uniformly from the interval $[0,1]$. Subsequently, each token in $\vx_0$ is replaced by \text{[M]} with probability $t$, yielding the corrupted sentence $\vx_t$. In the reverse process, LLaDA commences with a sentence composed entirely of \text{[M]} tokens and iteratively predicts these masked tokens to reconstruct the original sentence. We provide detailed formulations and sampling processes of masked diffusion models in Appendix~\ref{app:formulation_mask_diffusion_models}.

\textbf{Visual Instruction Tuning}~\cite{liu2023visual,liu2024improved,li2024llava} is a mainstream Multimodal Large Language Model (MLLM) architecture, recognized for its powerful performance and data efficiency. Specifically, it comprises a vision tower (e.g., CLIP~\cite{radford2021learning} or SigLIP~\cite{zhai2023sigmoid,tschannen2025siglip}) that converts images into visual representations, an MLP connector that projects these representations into an LLM's word embedding space, and the LLM itself. Through visual instruction tuning, this setup enables LLMs to achieve strong multimodal understanding capabilities with less than 1M image-text pairs.

\section{Method}
In contrast to predominant approaches that rely on autoregressive language models~\cite{liu2023visual, liu2024improved, li2024llava, chen2024internvl, wang2024qwen2, team2024chameleon}, our research explores how to perform visual instruction tuning~\cite{liu2023visual} in language diffusion models~\cite{nie2025large} for multimodal understanding. To this end, we formulate a training objective for multi-turn multimodal dialogues and explore the attention mechanism architectures (Sec.~\ref{sec:training_objective_architecture}), detail the inference process (Sec.~\ref{sec:inference_process_lladav}), and design a multi-stage training strategy (Sec.~\ref{sec:training_strategies_lladav}). These components collectively enable diffusion language models to effectively process multimodal inputs.

\subsection{Training Objective and Architecture}
\label{sec:training_objective_architecture}
As with most MLLMs, the training of LLaDA-V utilizes multimodal understanding data involving multi-turn dialogues. For simplicity, we use a sample consisting of a single image and a two-turn dialogue as an example. As LLaDA-V represents an early exploration into applying large language diffusion models for multimodal understanding, its design prioritizes simplicity, effectiveness, and alignment with established training methodologies of autoregressive-based MLLMs. Consequently, we adopt the seminar visual instruction tuning framework~\cite{li2024llava}, comprising a language tower, a vision tower, and an MLP projector. For the language tower, we selected LLaDA~\cite{nie2025large}, a representative large language diffusion model with language performance comparable to LLaMA3-8B, enabling us to explore the capabilities of purely diffusion-based MLLMs. For the vision tower and MLP projector, we selected SigLIP 2~\cite{tschannen2025siglip} and a two-layer MLP, respectively, due to their demonstrated effectiveness across various MLLMs. 

For training the aforementioned models within LLaDA-V, we now present the necessary notations and training objective. Let $\vv$ denote the image representation from the vision tower and MLP projector and \text{[M]} denote a special mask token. For a two-turn dialogue, we denote the data instance as $(\vv, \vp_0^1, \vr_0^1, \vp_0^2, \vr_0^2)$, where $\vp_0^{1} = [\vp_0^{1, i}]_{i=1}^{L_{p1}}$ and $\vp_0^{2} = [\vp_0^{2, i}]_{i=1}^{L_{p2}}$ are the prompts for the first and second turns, while $\vr_0^{1} = [\vr_0^{1, i}]_{i=1}^{L_{r1}}$ and $\vr_0^{2} = [\vr_0^{2, i}]_{i=1}^{L_{r2}}$ are their corresponding ground-truth responses. 

Formally, the training objective for LLaDA-V, $\mathcal{L}(\vtheta)$, is defined as:
\begin{align}
    \label{eq:llada-v-objective}
    - \mathbb{E}_{\substack{\vv, t, \vp_0^1, \vr_0^1, \vr_t^1,\\ \vp_0^2, \vr_0^2, \vr_t^2}} \left[\frac{1}{t} \sum_{i=1}^{L_{r1}} \sum_{j=1}^{L_{r2}} \textbf{1}[\vr_t^{1,i} = \text{[M]} \land \vr_t^{2,j} = \text{[M]}] \log p_{\vtheta}(\vr_0^{1,i},\vr_0^{2,j}|\vv, \vp_0^1, \vr_t^1, \vp_0^2, \vr_t^2) \right],
\end{align}
where $\vr_t^1$ and $\vr_t^2$ denote the masked response. 

\begin{figure}[t!]
    \centering
    \includegraphics[width=\textwidth]{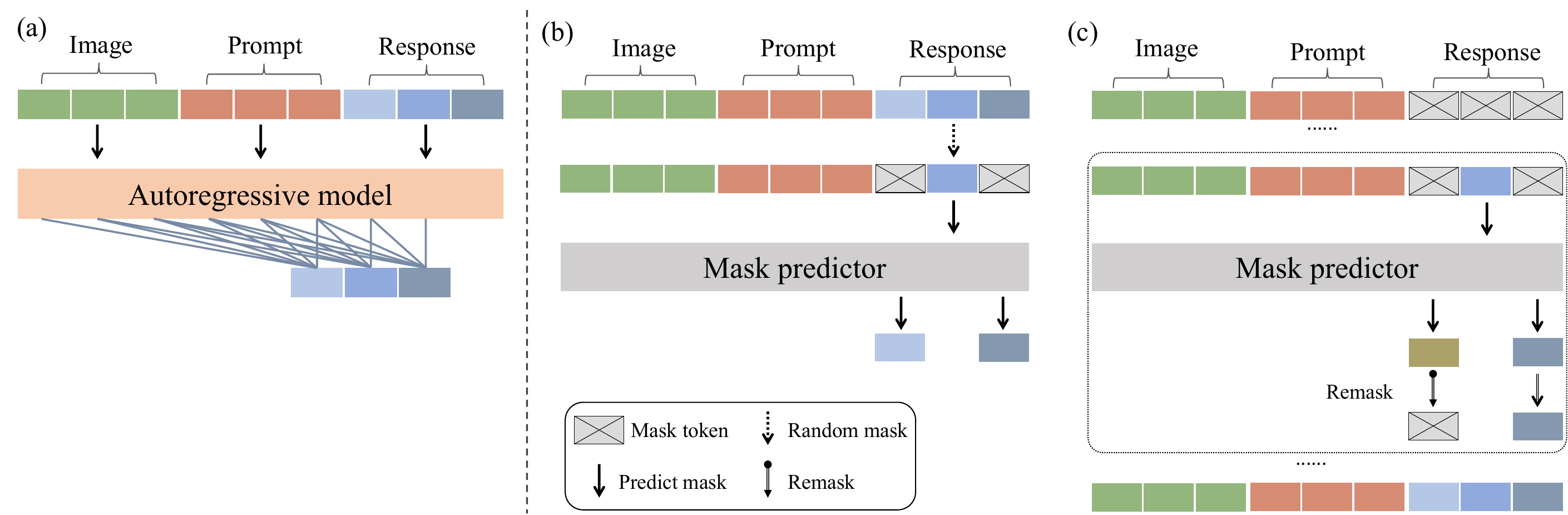}
    \caption{\textbf{Overview of Autoregressive Approaches and LLaDA-V.} Image representations are generated by an encoder and an MLP projector (not explicitly shown). (a) Autoregressive Training: Given image features and the input prompt, autoregressive models are trained to predict the response through next-token prediction. (b) LLaDA-V's Training: Image features and the input prompt remain unmasked, while only the response is randomly masked. (c) LLaDA-V's Inference: As time step $t$ decreases from 1 to 0, generation begins with a fully masked response and iteratively predicts tokens.}
    \label{fig:llada_v_architecture}
\end{figure}

Theoretically, the training objective in Eq.~\eqref{eq:llada-v-objective} has been proven to be an upper bound of the negative log-likelihood for masked tokens~\cite{ou2024your,shi2024simplified}. Intuitively, as shown in Fig.~\ref{fig:llada_v_architecture} (b), the training objective aims to predict masked tokens within the response, given clean image features and prompts. Through Eq.~\eqref{eq:llada-v-objective} and visual instruction tuning framework~\cite{liu2023visual}, we effectively extend the large language diffusion model to encompass multimodal understanding capabilities. 

Regarding the architecture of LLaDA-V, our primary focus is on exploring the attention mechanism design within the language tower. To mitigate the potential gap between training and inference (see Sec.~\ref{sec:inference_process_lladav}), one might expect to use a causal attention structure during training for multi-turn dialogues (i.e., preventing an earlier turn like $\vp_0^1, \vr_0^1$ from accessing a later turn such as $\vp_0^2, \vr_0^2$). However, a bidirectional attention mechanism enables comprehensive understanding of the entire dialogue context during mask prediction, which has demonstrated its effectiveness in recent video diffusion models~\cite{wang2025wan, yang2024cogvideox, kong2024hunyuanvideo} where it enhances temporal consistency of generated video. Therefore, we conduct ablation studies on these two attention mechanism choices in Sec.~\ref{sec:ablation_study}, and observe that the bidirectional attention mechanism achieves superior results across more benchmarks. Based on these findings, we adopt the bidirectional attention mechanism in LLaDA-V.

\subsection{Inference Process}
\label{sec:inference_process_lladav}

Once the model is trained with the objective in Eq.~\eqref{eq:llada-v-objective}, LLaDA-V can generate multi-turn dialogues through iterative response generation. When given a new prompt, the model leverages previous prompts and responses to generate appropriate subsequent responses. While dialogue generation proceeds turn by turn, LLaDA-V differs by generating each response via the reverse process of a masked diffusion model, rather than next-token prediction used in autoregressive models. 

As shown in Fig.~\ref{fig:llada_v_architecture} (c), we illustrate the inference process using a one-turn dialogue example. Following this process, we generate samples from the distribution $p_{\vtheta}(\vr_0|\vv, \vp_0)$ by initializing with a fully masked response $\vr_1$ and applying the reverse process of the masked diffusion model, as detailed in Appendix~\ref{app:formulation_mask_diffusion_models}. Sampling starts by setting a target generation length and initializing the response $\vr_1$ entirely with \text{[M]} tokens. The sequence is iteratively refined by transitioning from a state $\vr_t$ to a state $\vr_s$ (representing decreasing mask levels, with $s<t$). Each step involves two main phases: first, LLaDA-V, conditioned on $\vv, \vp_0, \text{and } \vr_t$, predicts all \text{[M]} tokens in $\vr_t$. Second, to form $\vr_s$, a fraction $s/t$ of these predictions are re-masked to \text{[M]}, while the remainder ($1-s/t$) are kept as predicted, consistent with the reverse process of masked diffusion models. For the remasking strategy, rather than using standard random selection, we primarily adopt LLaDA's~\cite{nie2025large} \emph{low-confidence} strategy, which preferentially re-masks low-confidence predictions while preserving high-confidence ones. We choose this approach based on its consistently demonstrated improvements across various tasks.

\subsection{Training Strategies}
\label{sec:training_strategies_lladav}
We adopt a multi-stage training paradigm for LLaDA-V, with the first two stages following established practices in MLLMs like LLaVA-NeXT~\cite{liu2024llavanext} to establish language-vision alignment and build visual instruction following abilities. We further enhance this paradigm with a third stage focused on multimodal reasoning, enabling comprehensive capabilities across diverse tasks.

\textbf{Stage 1: Language-Image Alignment.} In this stage, we train the MLP projector to align visual representations with LLaDA's word embeddings, following established MLLM practices~\cite{liu2024improved, li2024llava, guo2024mammoth}. The language and vision towers remain frozen throughout this process. We utilize the LLaVA-Pretrain dataset~\cite{liu2023visual} for this alignment stage.

\textbf{Stage 2: Visual Instruction Tuning.}
Following language-image alignment, Stage 2 focuses on developing LLaDA-V's comprehensive multimodal understanding capabilities by fine-tuning the entire model on large-scale instruction data. This fine-tuning, which utilizes high-quality, large-scale multimodal instruction data from MAmmoTH-VL~\cite{guo2024mammoth}, aims to establish strong visual instruction-following abilities and enable the model to handle diverse scenarios involving single images, multiple images, or video inputs. Stage 2 is conducted in two distinct phases as follows. 

\begin{itemize}[leftmargin=*]
\setlength\itemindent{1em}
    \item \emph{Single Image Training}: The model is trained on 10M single-image multimodal data to establish image understanding capabilities. In this phase, LLaDA-V develops strong performance in recognizing and interpreting single images to respond to diverse instructions.
    \item \emph{OneVision Training}: Following single-image training, the model is further trained on approximately 2M diverse multimodal samples (single-image, multi-image, and video data). This phase expands LLaDA-V's capabilities to handle complex scenarios involving multiple images and temporal information beyond single-image contexts.
\end{itemize}

\textbf{Stage 3: Multimodal Reasoning Enhancement.}
Following visual instruction-following, Stage 3 focuses on enhancing multimodal reasoning capabilities for complex tasks through two key steps:

\begin{itemize}[leftmargin=*]
\setlength\itemindent{1em}
    \item \emph{Reasoning Training}: In this step, we trained LLaDA-V on reasoning-focused multimodal data from VisualWebInstruct~\cite{jia2025visualwebinstruct}, which contains 900K QA pairs featuring detailed reasoning chains and final answers. This training phase is designed to enhance the model's ability to perform complex multimodal reasoning.
    \item \emph{Balanced Reasoning Training}: Following \emph{reasoning training}, LLaDA-V consistently provided explicit reasoning before answers. To enhance response flexibility, a subsequent phase, inspired by Qwen 3's hybrid thinking mechanism~\cite{qwen3blog}, utilized a mixed dataset: reasoning-focused VisualWebInstruct combined with MAmmoTH-VL's OneVision data. In this mixed training, `/no\_think' tags were appended to OneVision prompts to encourage direct answers, while `/think' tags were applied to 50\% of reasoning-data prompts.
\end{itemize}

\section{Experiment}
\label{sec:exp}
This section presents our experimental setup and results, including: experimental settings (Sec.~\ref{sec:exp_settings}); data scaling experiments (Sec.~\ref{sec:data_scalability}); comprehensive benchmark evaluations (Sec.~\ref{sec:benchmark_results}); and ablation studies on attention mask selection (Sec.~\ref{sec:ablation_study}).

\begin{table}[t!]
    \centering
    \caption{\textbf{Training Settings.} Here M-SI and M-OV represent the single image data and onevision data of MAmmoTH~\cite{guo2024mammoth}, while VW represents the data of VisualWebInstruct~\cite{jia2025visualwebinstruct}. We train LLaDA-V sequentially through the first five datasets (LLaVA-Pretrain~\cite{liu2023visual}, M-SI, M-OV, VW, and M-OV+VW), while the last dataset (LLaVA-NeXT~\cite{liu2024llavanext}) is used for ablation study in Sec.~\ref{sec:ablation_study}.}
    \label{tab:llada_training_hyperparameters}
    \vspace{.2cm}
    \begin{adjustbox}{max width=\textwidth}
    \begin{tabular}{l| c | c c | c c | c}
      \toprule
      Training data & LLaVA-Pretrain & M-SI & M-OV & VW & M-OV+VW & LLaVA-NeXT \\
      \midrule
      Vision tower & \multicolumn{6}{c}{Siglip2-so400m-patch14-384~\cite{tschannen2025siglip}} \\
      Language tower & \multicolumn{6}{c}{LLaDA-8B-Instruct~\cite{nie2025large}} \\
      Attention &  \multicolumn{6}{c}{Bidirectional attention} \\
      \midrule 
      Batch size & 64 & 256 & 256 & 256 & 256 & 64 \\
      Model max length & 8192 & 8192 & 16384 & 8192 & 16384 & 8192\\
      \#Samples & 558K & 10M & 2M & 900K & 3M & 738K \\
      \midrule
      LR of vision tower & - & \multicolumn{2}{c|}{$2\times 10^{-6}$} & \multicolumn{2}{c|}{$2\times 10^{-6}$} & $2\times 10^{-6}$ \\
      LR of language tower & - & \multicolumn{2}{c|}{$1\times 10^{-5}$} & \multicolumn{2}{c|}{$1\times 10^{-5}$} & $1\times 10^{-5}$ \\
      LR of projector & $1\times 10^{-3}$ & \multicolumn{2}{c|}{$1\times 10^{-5}$} & \multicolumn{2}{c|}{$1\times 10^{-5}$} & $1\times 10^{-5}$\\
      Epoch & 1 & \multicolumn{2}{c|}{1} & \multicolumn{2}{c|}{1} & 1 \\
      \bottomrule
    \end{tabular}
    \end{adjustbox}
\end{table}

\subsection{Experimental Settings}
\label{sec:exp_settings}
\textbf{Model.} We use LLaDA-8B-Instruct~\cite{nie2025large} for the language tower of LLaDA-V, an open-source diffusion-based large language model with extensive pre-training and supervised fine-tuning (SFT). However, it lacks preference alignment techniques~\cite{schulman2017proximal, rafailov2023direct, meng2024simpo, shao2024deepseekmath} that enhance conversational and reasoning capabilities in contemporary LLMs~\cite{touvron2023llama2, grattafiori2024llama}. Consequently, its performance falls behind Qwen2.5-7B-Instruct~\cite{yang2024qwen2} and is marginally inferior to LLaMA3-8B-Instruct~\cite{grattafiori2024llama}. For a fair comparison between LLaDA-V and autoregressive approaches, we use LLaMA3-8B-Instruct as the language tower in our primary baseline model, while maintaining all other components identical to LLaDA-V. For the vision tower, we utilize siglip2-so400m-patch14-384~\cite{tschannen2025siglip}, which offers robust visual representation capabilities. The projector is implemented as a randomly initialized two-layer MLP.

\textbf{Data.} For Stage 1, we employ the alignment dataset from LLaVA-Pretrain~\cite{liu2023visual}. In Stage 2, we leverage the comprehensive MAmmoTH-VL~\cite{guo2024mammoth} dataset, which consists of two primary components: SI-10M, comprising 10 million single-image multimodal samples, and OV-2M, containing 2 million diverse samples across single-image, multi-image, and video modalities. For Stage 3, we utilize the reasoning-focused multimodal dataset VisualWebInstruct~\cite{jia2025visualwebinstruct}. To achieve balanced reasoning capabilities, we further incorporate OV-2M into this stage of training. A comprehensive description of these training strategies can be found in Section~\ref{sec:training_strategies_lladav}.

\textbf{Training.} As detailed in Sec.~\ref{sec:training_strategies_lladav}, the LLaDA-V training process consists of three stages. In the first stage, only the Projector is trained. Subsequently, the full model is trained during the second and third stages. Detailed training settings can be found in Tab.~\ref{tab:llada_training_hyperparameters}.

\textbf{Evaluation.} 
To comprehensively evaluate LLaDA-V's performance, we considered multiple vision-language benchmarks across several categories:

\begin{itemize}[leftmargin=*]
    \setlength\itemindent{1em}
    \item \emph{Multidisciplinary Knowledge \& Mathematical Reasoning:} 
    MMMU~\cite{yue2024mmmu}, MMMU-Pro~\cite{yue2024mmmupro}, MMStar~\cite{chen2024we}, MME~\cite{fu2023mme}, SeedBench~\cite{li2023seed}, MMBench~\cite{liu2024mmbench}, MathVerse~\cite{zhang2024mathverse}, and MathVista~\cite{lu2023mathvista}.
    
    \item \emph{Chart \& Doc Understanding:} 
    AI2D~\cite{kembhavi2016diagram}, ChartQA~\cite{masry2022chartqa}, DocVQA~\cite{mathew2021docvqa}, and InfoVQA~\cite{mathew2022infographicvqa}.
    
    \item \emph{Real-world Scene Understanding:} 
    RealworldQA~\cite{x2024realworldqa}.
    
    \item \emph{Multi-image \& Video Understanding:} 
    MuirBench~\cite{wang2024muirbench}, MLVU~\cite{zhou2024mlvu}, and VideoMME~\cite{fu2024video}.
\end{itemize}

\subsection{Data Scalability of LLaDA-V}
\label{sec:data_scalability}
In order to demonstrate the effectiveness of LLaDA-V, we first evaluate the data scalability of LLaDA-V in comparison with the autoregressive baseline LLaMA3-V. To ensure a fair comparison between LLaDA-V and LLaMA3-V, we adopted an identical training pipeline for both models. The training process consisted of two main phases: first, we pretrained the projectors using LLaVA-Pretrain data~\cite{liu2023visual}; then, we conducted full model training (including vision tower, language tower, and projector) on the single-image data of MAmmoTH-VL~\cite{guo2024mammoth}. We evaluated the models' performance at various data scales using six carefully selected multimodal benchmarks. 

\begin{figure}[t!]
    \centering
    \begin{subfigure}[b]{0.325\textwidth}
      \centering
      \includegraphics[width=\textwidth]{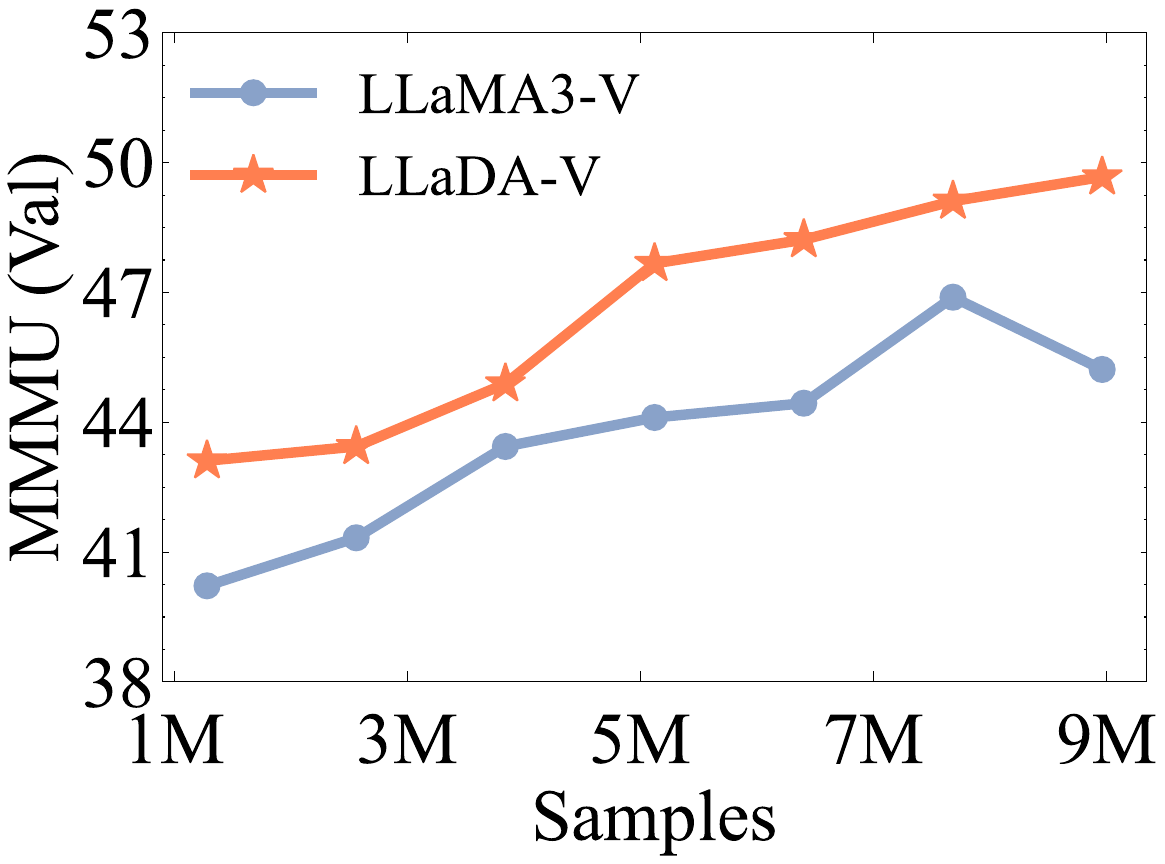}
    \end{subfigure}
    \begin{subfigure}[b]{0.325\textwidth}
      \centering
      \includegraphics[width=\textwidth]{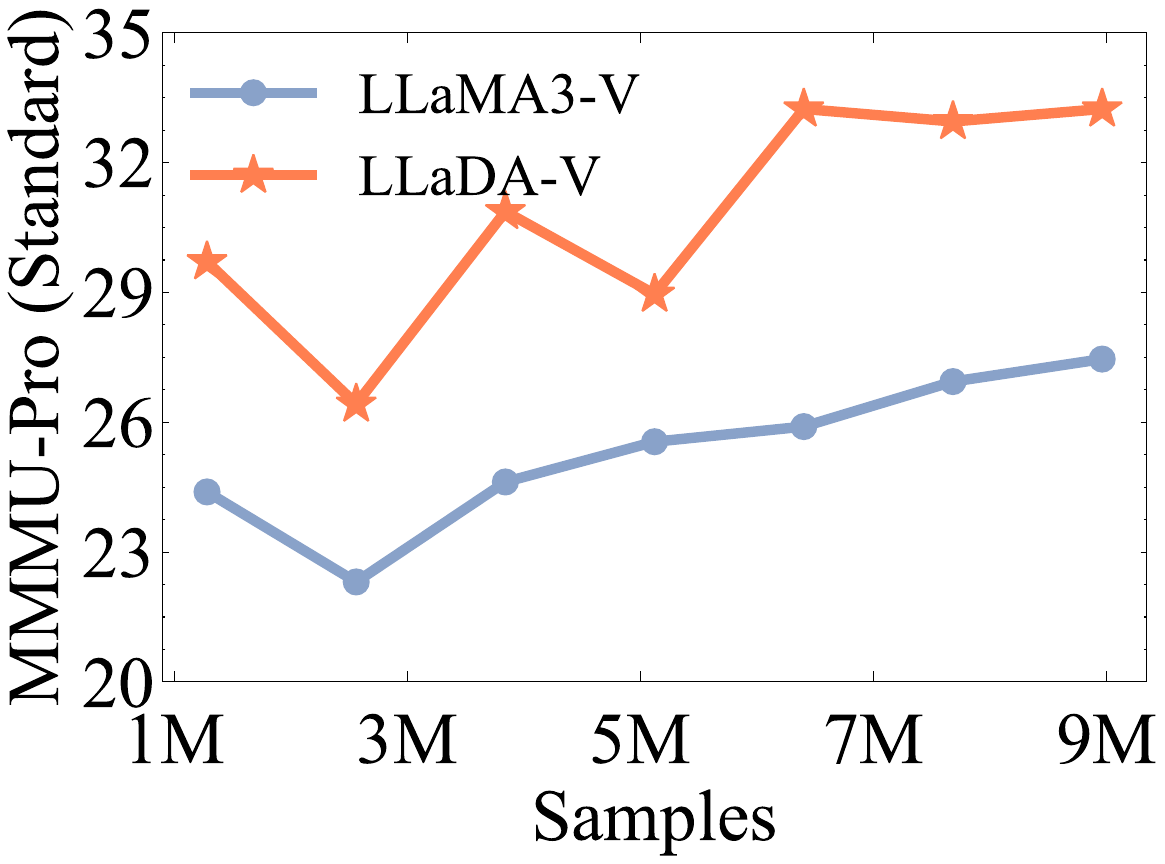}
    \end{subfigure}
    \begin{subfigure}[b]{0.325\textwidth}
      \centering
      \includegraphics[width=\textwidth]{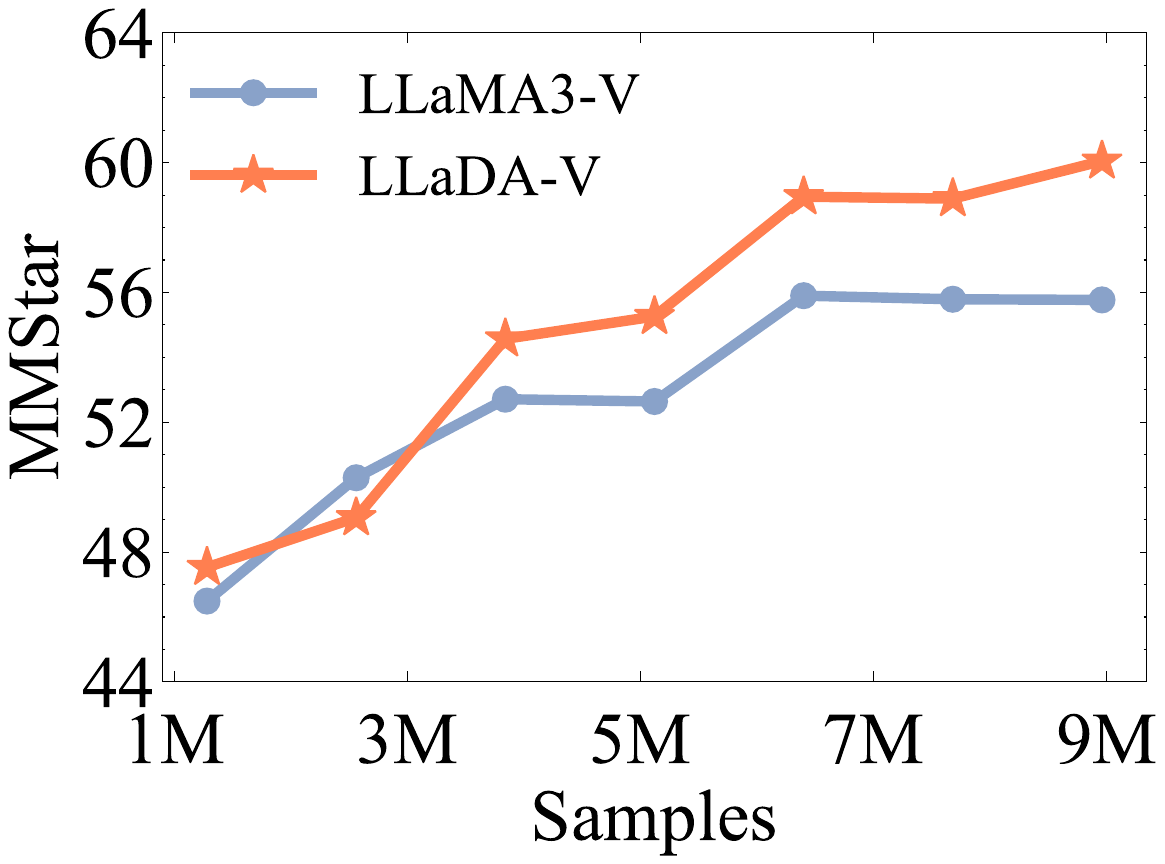}
    \end{subfigure} \\ 
    \begin{subfigure}[b]{0.325\textwidth}
        \centering
        \includegraphics[width=\textwidth]{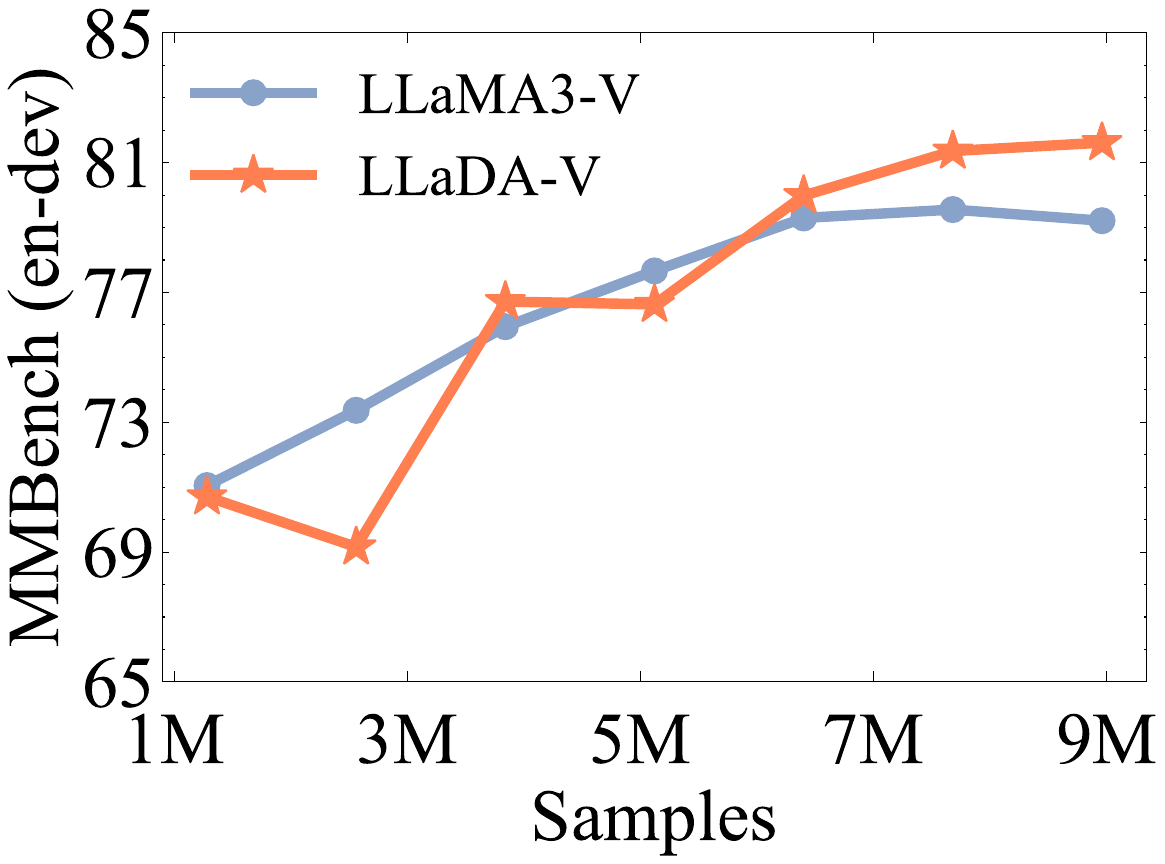}
      \end{subfigure}
    \begin{subfigure}[b]{0.325\textwidth}
      \centering
      \includegraphics[width=\textwidth]{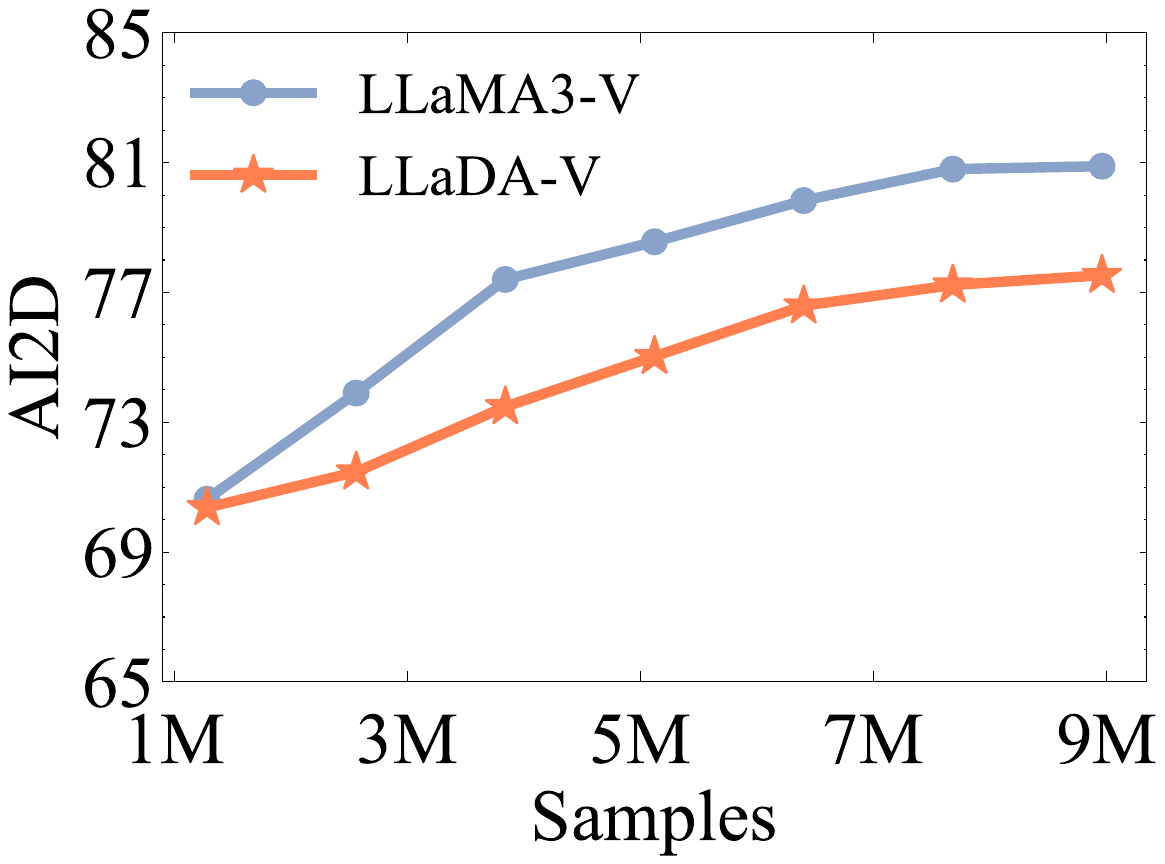}
    \end{subfigure}
    \begin{subfigure}[b]{0.325\textwidth}
      \centering
      \includegraphics[width=\textwidth]{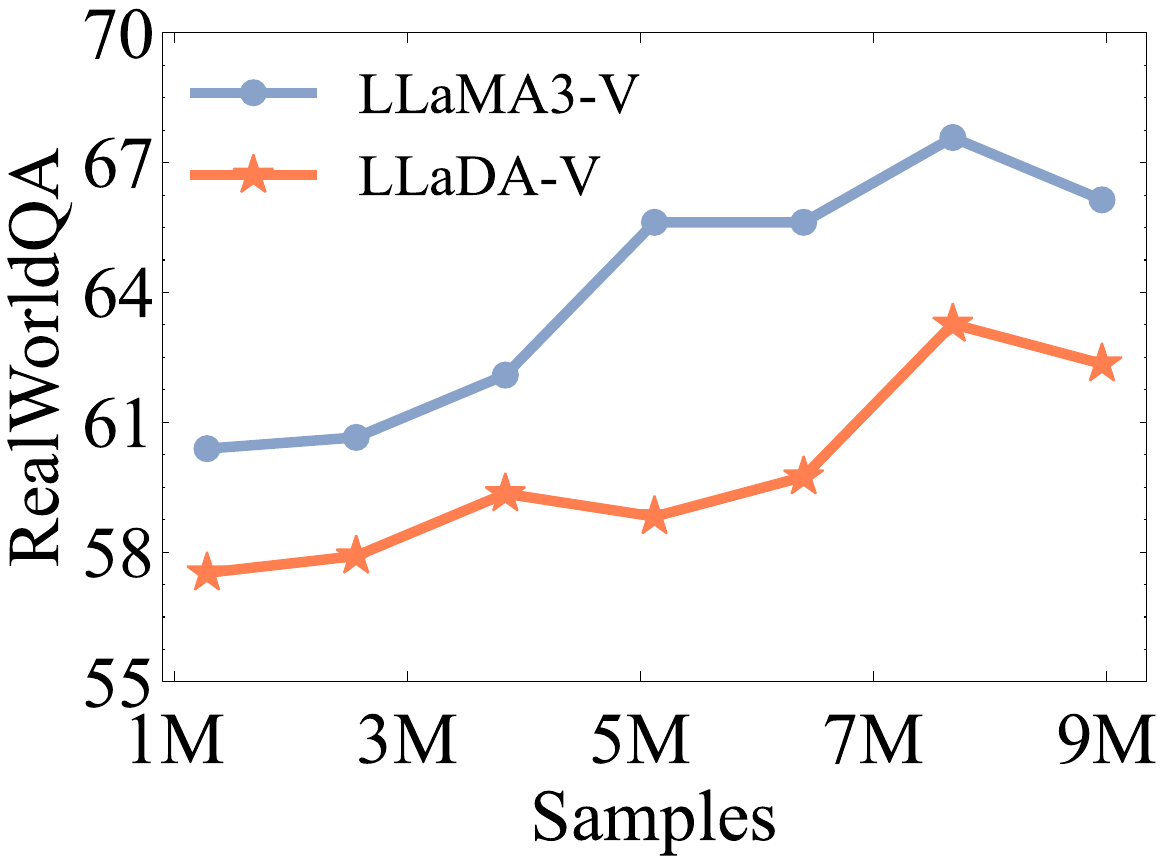}
    \end{subfigure} 
    \caption{\textbf{Data Scalability of LLaDA-V.} Both LLaDA-V and LLaMA3-V were trained on MAmmoTH-VL-SI10M, with performance evaluated across six multimodal benchmarks. Despite having a weaker language tower, LLaDA-V shows superior data scalability across more tasks, especially excelling in multidisciplinary knowledge and mathematical reasoning.
    }
    \label{fig:data_scaling_lladav}
  \end{figure}

As shown in Fig.~\ref{fig:data_scaling_lladav}, we observe two key findings: First, LLaDA-V's performance consistently improves with increasing training data, demonstrating that LLaDA-V benefits from data scalability. Second, despite using a slightly weaker language tower, LLaDA-V shows superior scalability compared to LLaMA3-V on multidisciplinary knowledge benchmarks such as MMMU~\cite{yue2024mmmu} and MMMU-Pro~\cite{yue2024mmmupro}. Notably, for MMMU-Pro, LLaDA-V trained with merely 1M samples outperforms LLaMA3-V trained with 9M samples. However, on benchmarks assessing chart/document understanding (e.g., AI2D) and real-world scene understanding (e.g., RealworldQA), LLaDA-V lags behind LLaMA3-V.

  \begin{table}[t!]
    \centering
    \caption{\textbf{Benchmark Results for Multidisciplinary Knowledge and Mathematical Reasoning Tasks.} ``Diffusion'' here encompasses both continuous and discrete diffusion models. \emph{Notably, LLaDA-V outperforms all other hybrid and pure diffusion MLLMs, surpassing LLaMA3-V on 6 of 9 benchmarks despite having a relatively weaker language tower.} For comparison, we list each model's language tower, as this significantly impacts MLLM performance. ``-'' indicates unavailable data.}
    \label{tab:llada_multi_reasoning}
    \vspace{.2cm}
    \begin{adjustbox}{max width=\textwidth}
    \setlength{\tabcolsep}{3pt} 
    \begin{tabular}{lcc|ccccccccc}
      \toprule
      Model & Type & LLM Tower & MMMU & MMMU-Pro & MMMU-Pro & MMStar & MME& SeedB & MMB  & MathVerse & MathVista \\  
      & & & val & standard & vision & test & cog./perp. & image & en-dev & mini-vision & testmini \\
      \midrule 
      ShareGPT4V\cite{chen2024sharegpt4v} & AR & Vicuna-7B & - & - & - & - & 376/1567 & 69.7 & 68.8 & - & - \\
      Cambrian-1\cite{tong2024cambrian} & AR & LLaMA3-8B & 42.7 & - & - & - & -/1547 & 74.7 & 75.9 & - & 49.0 \\
      LLaVA\cite{liu2023visual} & AR & Vicuna-7B & - & - & - & - & -/809 & 37.0 & 38.7 & - & - \\
      LLaVA-1.5\cite{liu2024improved}  & AR & Vicuna-7B & - & - & - & - & -/1510 & 66.1 & 64.3 & - & - \\
      Qwen2-VL\cite{wang2024qwen2} & AR & Qwen2-7B & 54.1 & 43.5 & - & 60.7 & - & - & - & - & 58.2 \\
      DeepSeek-VL\cite{lu2024deepseek} & AR & DeepSeek-7B & 36.6 & - & - & - & - & 70.4 & 73.2 & - & - \\
      DeepSeek-VL2\cite{wu2024deepseek} & AR & - & 51.1 & - & - & 61.3 & - & - & - & - & 62.8 \\
      Janus\cite{wu2024janus} & AR & DeepSeek-1.3B & 30.5 & - & - & - & -/1338 & 63.7 & 69.4 & - & - \\
      Janus-Pro\cite{chen2025janus} & AR & DeepSeek-7B & 41.0 & - & - & - & -/1567 & 72.1 & 79.2 & - & - \\
      Emu3\cite{wang2024emu3} & AR & - & 31.6 & - & - & - & - & 68.2 & 58.5 & - & - \\
      MAmmoTH\cite{guo2024mammoth} & AR & Qwen2.5-7B & 50.8 & - & 25.3 & 63.0 & - & 76.0 & - & 34.2 & 67.6 \\
      LLaVA-OV\cite{li2024llava} & AR & Qwen2-7B & 48.8 & - & - & 61.7 & 418/1580 & 75.4 & 80.8 & 26.2 & 63.2 \\
      \midrule
      MetaMorph\cite{tong2024metamorph} & AR+Diff. & LLaMA3.1-8B & 41.8 & - & - & - & - & 71.8 & 75.2 & - & -\\
      Show-o\cite{xie2024show} & AR+Diff. & Phi1.5-1.3B & 27.4 & - & - & - & -/1232 & - & - & - & -\\
      JanusFlow\cite{ma2024janusflow} & AR+Diff. & DeepSeek-1.3B & 29.3 & - & - & - & -/1333 & 70.5 & 74.9 & - & - \\
      Orthus\cite{kou2024orthus} & AR+Diff. & Chameleon-7B & 28.2 & - & - & - & -/1265 & - & - & - & - \\
      \midrule
      D-DiT\cite{li2024dual} & Diff. & - & - & - & - & - & -/1124 & - & - & - & - \\
      \midrule
      LLaMA3-V & AR & LLaMA3-8B & 45.4 & 28.3 & 14.5 & 56.5 & 446/1581 & \textbf{76.6} & 79.8 & \textbf{29.0} & \textbf{62.1} \\
      LLaDA-V & Diff. & LLaDA-8B & \textbf{48.6} & \textbf{35.2} & \textbf{18.6} & \textbf{60.1} & \textbf{491/1507} & 74.8 & \textbf{82.9} & 28.5 & 59.7 \\
      \bottomrule
    \end{tabular}
    \end{adjustbox}
\end{table}

\subsection{Benchmark Results}
\label{sec:benchmark_results}

To comprehensively assess LLaDA-V's multimodal understanding capabilities, we evaluated it against three different model architectures—autoregressive, hybrid autoregressive-diffusion, and pure diffusion models—across a diverse set of 18 benchmarks (detailed results in Tab.~\ref{tab:llada_multi_reasoning} and Tab.~\ref{tab:llada_ocr_video}). These benchmarks encompass areas such as multidisciplinary knowledge, mathematical reasoning, chart/document understanding, real-world scene understanding, and multi-image/video tasks. 

Notably, in these comparative evaluations, LLaDA-V consistently demonstrates superior performance among hybrid autoregressive-diffusion and pure diffusion models, such as MetaMorph~\cite{tong2024metamorph} and D-DiT~\cite{li2024dual}. Furthermore, when compared with our autoregressive baseline LLaMA3-V, LLaDA-V exhibited strengths in some tasks: it outperformed LLaMA3-V on most multidisciplinary knowledge and mathematical reasoning benchmarks (e.g. MMMU, MMMU-Pro), while also achieving superior performance in multi-image and video understanding tasks (e.g., MuirBench, MLVU). These results are impressive considering LLaDA-V uses a relatively weaker language tower (see results in Tab.2 of \cite{nie2025large}). However, its performance remained less competitive on tasks focused on chart/document understanding (e.g., AI2D, DocVQA) and real-world scene comprehension (e.g., RealworldQA). For a fair comparison, our autoregressive baseline LLaMA3-V shares identical training protocols with LLaDA-V (see Sec.~\ref{sec:training_strategies_lladav}), with the only difference being the language tower. 

When compared with the strong autoregressive-based MLLM Qwen2-VL~\cite{wang2024qwen2}, LLaDA-V generally underperforms across most benchmarks, only achieving comparable results on a limited number of specific tasks such as MMStar. The performance difference primarily stems from LLaDA-V's weaker language backbone (LLaDA-8B) compared to Qwen2-VL's Qwen2-7B (see results in Tab.2 of \cite{nie2025large}), since the language model's perfomance is crucial for MLLM's performance~\cite{laurenccon2024matters}. However, as language diffusion models continue to improve, diffusion-based MLLMs are expected to achieve better performance, gradually narrowing the gap with leading models such as Qwen2-VL.

\begin{table}[t!]
    \centering
    \caption{\textbf{Benchmark Results for Chart, Document, Real-world Scene, Multi-image, and Video Tasks.} ``Diffusion'' here encompasses both continuous and discrete diffusion models. \emph{Compared to LLaMA3-V, LLaDA-V shows comparable performance on chart/document tasks, performs less well on real-world scenes, but excels in multi-image and video tasks.} ``-'' indicates missing data.}
    \label{tab:llada_ocr_video}
    \vspace{.2cm}
    \begin{adjustbox}{max width=\textwidth}
    \setlength{\tabcolsep}{3pt} 
    \begin{tabular}{lcc|ccccccccc}
      \toprule
      Model & Type & LLM Tower & AI2D & ChartQA & DocVQA & InfoVQA & RealworldQA & SeedB & MuirBench  & MLVU & VideoMME \\  
      & & &  &  & val & val &  & video & & dev &  \\
      \midrule 
      Cambrian-1\cite{tong2024cambrian} & AR & LLaMA3-8B & 73.0 & 73.3 & - & - & 64.2 & - & - & - & - \\
      LLaVA\cite{liu2023visual} & AR & Vicuna-7B & - & - & - & - & - & 23.8 & - & - & - \\
      LLaVA-1.5\cite{liu2024improved}  & AR & Vicuna-7B & - & - & - & - & - & 37.3 & - & - & - \\
      Qwen2-VL\cite{wang2024qwen2} & AR & Qwen2-7B & 83.0 & 83.0 & - & - & 70.1 & - & - & - & - \\
      DeepSeek-VL2\cite{wu2024deepseek} & AR & - & 81.4 & 86.0 & - & - & 68.4 & - & - & - & - \\
      Emu3\cite{wang2024emu3} & AR & - & 70.0 & 68.6 & - & - & 57.4 & - & - & - & - \\
      MAmmoTH\cite{guo2024mammoth} & AR & Qwen2.5-7B & 84.0 & 86.2 & - & - & 69.9 & 57.1 & 55.1 & 64.7 & 58.8 \\
      LLaVA-OV\cite{li2024llava} & AR & Qwen2-7B & 81.4 & 80.0 & - & - & 66.3 & 56.9 & 41.8 & 64.7 & 58.2 \\
      \midrule
      MetaMorph\cite{tong2024metamorph} & AR+Diff. & LLaMA3.1-8B & - & 37.1 & - & - & 58.3 & - & - & - & - \\
      JanusFlow\cite{ma2024janusflow} & AR+Diff. & DeepSeek-1.3B & - & 64.6 & - & - & - & - & - & - & - \\
      
      \midrule
      LLaMA3-V & AR & LLaMA3-8B & \textbf{81.1} & 77.8 & \textbf{86.2} & 58.9 & \textbf{66.0} & \textbf{55.0} & 47.4 & 57.5 & 55.8 \\
      LLaDA-V & Diff. & LLaDA-8B & 77.8 & \textbf{78.3} & 83.9 & \textbf{66.3} & 63.2 & 53.7 & \textbf{48.3} & \textbf{59.5} & \textbf{56.1} \\
      \bottomrule
    \end{tabular}
    \end{adjustbox}
\end{table}

\subsection{Ablation Study}
\label{sec:ablation_study}
We adopt the two-stage training paradigm of LLaVA-NeXT~\cite{liu2024llavanext} for our ablation study. First, we train the MLP projector on the LLaVA-Pretrain dataset~\cite{liu2023visual}, then further fine-tune the entire model on the LLaVA-NeXT dataset~\cite{liu2024llavanext}. Training hyperparameter details are provided in Tab.~\ref{tab:llada_training_hyperparameters}.

We consider two attention mask strategies: dialogue causal and no mask (i.e., bidirectional attention). In the dialogue causal approach, earlier dialogue turns cannot attend to later turns. Conversely, the no mask strategy employs bidirectional attention, allowing attention across all turns. Further details on these masking architectures are available in Appendix~\ref{app:details_of_experiment}. As shown in Tab.~\ref{tab:llada_ablation}, the no mask strategy achieves superior performance, outperforming on 7 of the 12 benchmarks. We hypothesize that its underlying bidirectional attention mechanism provides a more comprehensive understanding of the entire dialogue context, thus improving model performance. This bidirectional attention mechanism is also widely adopted in recent video diffusion models~\cite{wang2025wan, yang2024cogvideox, kong2024hunyuanvideo} to improve temporal consistency. We thus adopt the no mask strategy in LLaDA-V.

\begin{table}[t!]
    \centering
    \caption{\textbf{Ablation Studies on Attention Mask.} Comparison of LLaDA-8B using different attention masking strategies (dialogue causal vs. no mask) across 12 benchmarks. We adopt the no mask strategy in LLaDA-V as it shows slightly better performance on most benchmarks.}
    \label{tab:llada_ablation}
    \vspace{.2cm}
    \begin{adjustbox}{max width=\textwidth}
    \begin{tabular}{l|cc}
      \toprule
      LLM Backbone & LLaDA~\cite{nie2024scaling} & LLaDA~\cite{nie2024scaling} \\
      Attention Mask & Dialogue Causal Mask & No Mask \\
      \midrule
          MMMU~\cite{yue2024mmmu}{\small(val)} & 42.89 & \textbf{44.67} \\
          MMMU-Pro~\cite{yue2024mmmupro}{\small(standard)} & 26.01 & \textbf{26.59} \\
          MMMU-Pro~\cite{yue2024mmmupro}{\small(vision)} & 11.56 &  \textbf{11.68} \\
          MMStar~\cite{chen2024we}& 49.60 & \textbf{49.79} \\
          MME~\cite{fu2023mme}{\small(cog./perp.)} & \textbf{365/1412} & 352/1370 \\
          SeedBench~\cite{li2023seed}{\small(image)}  & \textbf{72.16} & 71.59 \\
          SeedBench~\cite{li2023seed}{\small(video)}  & \textbf{45.75} & 45.54 \\
          MMBench~\cite{liu2024mmbench}{\small(en-dev)} & 75.42 & \textbf{76.71} \\
          AI2D~\cite{kembhavi2016diagram}& 70.89 & \textbf{71.47} \\
          ChartQA~\cite{masry2022chartqa}& \textbf{55.20} & 54.88 \\
          RealworldQA~\cite{x2024realworldqa}& \textbf{61.18} & 60.26 \\
          MuirBench~\cite{wang2024muirbench}& 28.69 & \textbf{33.88} \\
      \bottomrule
    \end{tabular}
    \end{adjustbox}
\end{table}

\section{Related Work}

\textbf{Diffusion Language Models.} Recently, diffusion language models have attracted increasing attention, including both continuous~\citep{li2022diffusion,gong2022diffuseq,han2022ssd,strudel2022self,chen2022analog,dieleman2022continuous,richemond2022categorical,wu2023ardiffusion,mahabadi2024tess,ye2023dinoiser,zhang2023planner,lou2023reflected,graves2023bayesian,lin2023text,xue2024unifying, zhang2025target} and discrete~\citep{hoogeboom2021argmax,campbell2022continuous, he2022diffusionbert, hoogeboom2021autoregressive,meng2022concrete,reid2022diffuser,sun2022score,kitouni2023disk,Zheng2023ARD,chen2023fast,ye2023diffusion,gat2024discrete,zheng2024maskeddiffusionmodelssecretly} variants. Among them, the masked diffusion models, a subclass of discrete diffusion models, have achieved the best performance. \citet{ou2024your, shi2024simplified, shao2024deepseekmath} established the theoretical foundations of masked diffusion models and demonstrated their competitiveness with autoregressive models at the GPT-2 scale. LLaDA~\citep{nie2025large} scales masked diffusion models to 8B parameters, making it the first diffusion-based language model that can rival modern LLMs such as LLaMA3 across a wide range of downstream tasks. While LLaDA's language performance remains slightly inferior to LLaMA3-8B, LLaDA-V shows superior performance across more tasks compared to our LLaMA3 baseline. This suggests LLaDA-V's framework may offer inherent advantages for multimodal applications.

\textbf{Multimodal Understanding.} Multimodal Large Language Models (MLLMs) have made significant strides by integrating multiple input modalities with strong Large Language Models (LLMs)~\cite{radford2018improving,radford2019language, brown2020language, touvron2023llama, touvron2023llama2, grattafiori2024llama,yang2024qwen2,li2023textbooks,bi2024deepseek}. From the perspective of the probabilistic modeling methods, MLLMs are primarily classified into three categories: autoregressive models~\cite{liu2023visual, liu2024improved, li2024llava, chen2024internvl, wang2024qwen2, team2024chameleon,ding2025kimi, chu2023qwen, ghosh2024gama, wang2025internvideo2, chen2024sharegpt4video, zhang2024video}, autoregressive-diffusion hybrid models~\cite{bao2023one,xie2024show,zhou2024transfusion,ma2024janusflow}, and pure diffusion models~\cite{swerdlow2025unified,li2024dual}. The most closely related work, D-DiT~\cite{li2024dual}, combines continuous diffusion for visual content with discrete diffusion for text. However, its limited language modeling capacity results in performance that falls significantly behind autoregressive and hybrid approaches. In contrast, LLaDA-V leverages a powerful language diffusion model~\cite{nie2025large} with an effective training framework to achieve state-of-the-art results among both hybrid and purely diffusion-based MLLMs

\section{Conclusion}
\label{sec:conclusions}
We present \textbf{LLaDA-V}, a purely diffusion-based Multimodal Large Language Model (MLLM) for both training and sampling, which builds upon the visual instruction tuning framework~\cite{liu2023visual} and the large language diffusion model~\cite{nie2025large}. LLaDA-V demonstrates superior performance among hybrid autoregressive-diffusion and purely diffsion-based model. Besides, LLaDA-V achieve better data scalability and performance across more benchmarks than LLaMA3-V, which employs a different language tower but shares the same training strategy. We effectively extend the large language diffusion model to encompass multimodal understanding capabilities. 

\textbf{Limitations.} A limitation of our work is the image processing strategy. For high-resolution images, we split and resize image segments, process them through our SigLIP2~\cite{tschannen2025siglip} vision tower, and concatenate the features. Unlike Qwen2-VL with native dynamic resolution support, this approach may reduce efficiency and accuracy in visual representation. We leave the development of more advanced image processing strategies for future work. 

\textbf{Broader Impacts.} 
We believe that LLaDA-V can inspire further exploration of probabilistic modeling approaches for multimodal understanding. However, like many advanced Multimodal Large Language Models (MLLMs), LLaDA-V may generate hallucinations—factually incorrect content or information not present in the input. Nonetheless, approaches such as scaling up data and developing more advanced alignment techniques may help mitigate this problem. 

\bibliography{ref}
\bibliographystyle{IEEEtranN}



\appendix
\newpage

\begin{tcolorbox}[
    sharp corners,
    boxrule=0.8pt,
    colback=white,
    colframe=black,
]
    {\hypersetup{hidelinks}
    \tableofcontents
    }
\end{tcolorbox}

\newpage

\section{The Formulation of Masked Diffusion Models}
\label{app:formulation_mask_diffusion_models}
In this section, we present the main formulation of masked diffusion models for completeness. Please refer to~\citet{shi2024simplified, sahoo2024simple, ou2024your} for theoretical details.

In masked diffusion models, the forward process independently masks each token in a sentence $\vx_0 \in \{0, 1, \dots, K{-}1\}^N$, based on a given noise level $t \in [0,1]$, where $K$ and $N$ denote the vocabulary size and sentence length, respectively.
\begin{align}
    q_{t|0}(\vx_t|\vx_0) = \prod_{i=0}^{N-1} q_{t|0}(\vx_t^i|\vx_0^i) \quad \text{and} \quad 
    q_{t|0}(\vx_t^i|\vx_0^i) = \begin{cases}
                        \alpha_t, & \vx_t^i = \vx_0^i, \\
                        1 - \alpha_t, & \vx_t^i = \text{[M]}.
                        \end{cases}
\end{align}
In LLaDA-V, we choose $\alpha_t=1-t$ following LLaDA~\cite{nie2025large} due to its demonstrated superior empirical performance. Intuitively, during the forward process, each token independently has a probability $t$ of being masked (replaced with [M]) and a probability $1-t$ of remaining unchanged.

Masked diffusion models generate text by simulating a reverse process that gradually transforms masked tokens into meaningful content, starting from a fully masked sequence. Given $0\le s < t \le 1$, each sampling step in the reverse process is characterized by
\begin{align}
\label{eq:reverse}
    q_{s|t}(\vx_s|\vx_t) = \prod_{i=0}^{N-1} q_{s|t}(\vx_s^i|\vx_t)~~ \text{and} ~~
    q_{s|t}(\vx_s^i|\vx_t) = 
        \begin{cases}
            1, & \vx_t^i \neq \text{[M]}, \vx_s^i = \vx_t^i,\\
            \frac{1-\alpha_s}{1-\alpha_t}, & \vx_t^i = \text{[M]}, \vx_s^i = \text{[M]}, \\
            \frac{\alpha_s-\alpha_t}{1-\alpha_t} p_{\vtheta}(\vx_0^i|\vx_t), & \vx_t^i =\text{[M]}, \vx_s^i \neq \text{[M]}, \\
            0, & \textrm{otherwise},
        \end{cases}
\end{align}
where $p_{\vtheta}$ is modeled by a Transformer. When using $\alpha_t=1-t$, the reverse process has an intuitive interpretation: at each generation step, tokens that are already meaningful content remain unchanged, while masked tokens [M] either stay masked with probability $s/t$ or are replaced with meaningful content predicted by the model with probability $1-s/t$.

The training objective of masked diffusion models is the following upper bound on negative log-likelihood:
\begin{align}
    \mathcal{L}_{\vtheta} = \int_0^1 \frac{1}{t} \mathbb{E}_{q(\vx_t|\vx_0)}\left[ \sum_{\{i| \vx_t^i = m\}} - \log p_{\vtheta}(\vx_0^i|\vx_t) \right] d t.
\end{align}

For each sampling step in the reverse process (Eq.~\eqref{eq:reverse}), given $\vx_t$, we first identify masked positions $i$ (where $\vx_t^i = \text{[M]}$) and then sample a token $\vx_0^i$ for each such position from the distribution $p_{\theta}(\vx_0^i \mid \vx_t)$. Subsequently, a fraction $s/t$ of these newly sampled tokens are typically selected randomly for re-masking. However, \citet{chang2022maskgit} introduced a deterministic re-masking strategy that selects tokens with the lowest confidence scores (i.e., the smallest $p_{\theta}(\vx_0^i \mid \vx_t)$ values) for re-masking, comprising the $s/t$ proportion. LLaDA~\citep{nie2025large} adopts this low-confidence re-masking approach and demonstrates consistent improvements across various downstream tasks. In LLaDA-V, we also employ this low-confidence re-masking strategy following LLaDA.

\section{Experiments}
\label{app:details_of_experiment}
The implementation of LLaDA-V leverages official codebases and datasets from MAmmoTH~\cite{guo2024mammoth}, VisualWebInstruct~\cite{jia2025visualwebinstruct}, LLaVA-NeXT~\cite{liu2024llavanext}, and LMMS-EVAL~\cite{zhang2024lmmsevalrealitycheckevaluation}, with details of the corresponding links provided in Tab.~\ref{tab:code_link_license}.

\begin{table}[t]
    \centering
    \caption{\textbf{Code repositories and datasets leveraged in our implementation}}
    \label{tab:code_link_license}
    \vspace{.2em}
    \begin{adjustbox}{max width=\textwidth}
    \begin{tabular}{lc}
    \toprule
    \textbf{Code} & \textbf{URL} \\
    \midrule
        LMMs-Eval & \url{https://github.com/EvolvingLMMs-Lab/lmms-eval} \\ 
        LLaVA-NeXT & \url{https://github.com/LLaVA-VL/LLaVA-NeXT} \\ 
        MAmmoTH-VL & \url{https://github.com/MAmmoTH-VL/MAmmoTH-VL} \\ 
        VisualWebInstruct & \url{https://github.com/TIGER-AI-Lab/VisualWebInstruct} \\
    \midrule
        \textbf{Data}  & \textbf{URL} \\
    \midrule 
    LLaVA-Pretrain & \url{https://huggingface.co/datasets/liuhaotian/LLaVA-Pretrain} \\ 
    LLaVA-NeXT & \url{https://huggingface.co/datasets/lmms-lab/LLaVA-NeXT-Data} \\
    MAmmoTH-VL &  \url{https://huggingface.co/datasets/MAmmoTH-VL/MAmmoTH-VL-Instruct-12M} \\
    VisualWebInstruct & \url{https://huggingface.co/datasets/TIGER-Lab/VisualWebInstruct} \\ 
    \bottomrule
    \end{tabular}
    \end{adjustbox}
\end{table}

\subsection{Model Architecture}
The language tower of LLaDA-V strictly follows the architecture of LLaDA~\citep{nie2025large}. The architecture of LLaDA is largely based on LLaMA3~\citep{touvron2023llama}, with the main difference being the removal of the causal mask: LLaDA replaces the causal transformer in LLaMA3 with a bidirectional transformer. As a result, LLaDA does not support KV caching and uses standard multi-head attention, in contrast to the grouped query attention~\citep{ainslie2023gqa} in LLaMA3. Aside from these changes, both models employ widely used techniques in large language models, including RMSNorm~\citep{zhang2019root}, SwiGLU~\citep{shazeer2020glu}, and RoPE~\citep{su2024roformer}. For the vision tower in LLaDA-V, we employ the siglip2-so400m-patch14-384 model, which processes visual inputs with a resolution of 384$\times$384 pixels and produces 729 visual tokens per image. For the projector in LLaDA-V, we employ a randomly initialized two-layer MLP.

\subsection{Attention Mask}
\begin{figure}[t!]
    \centering
    \begin{subfigure}[b]{0.325\textwidth}
        \centering
        \includegraphics[width=\textwidth]{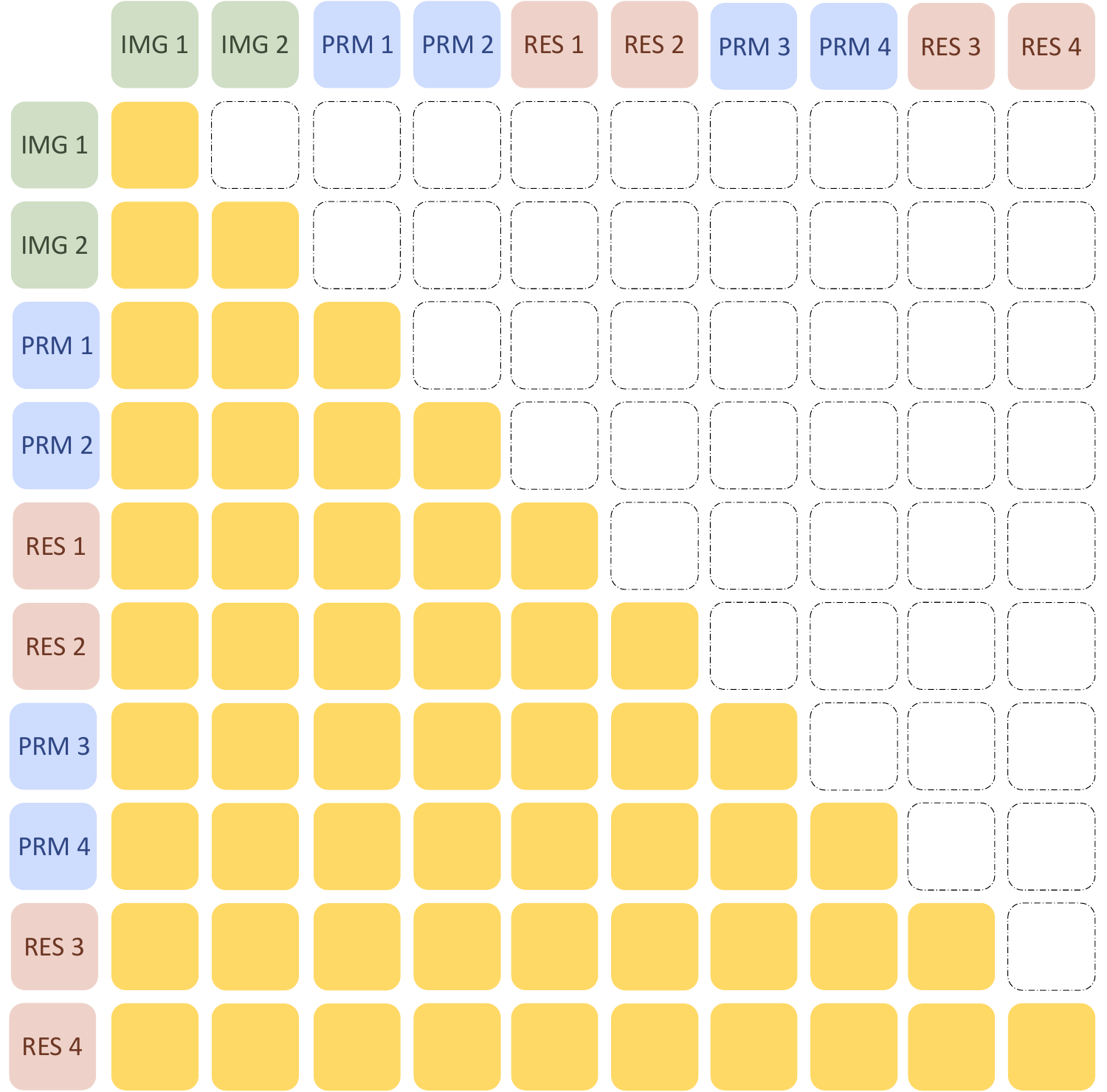}
        \caption{Causal Mask}
        \label{fig:ar_attn}
      \end{subfigure}
    \hfill
    \begin{subfigure}[b]{0.325\textwidth}
        \centering
        \includegraphics[width=\textwidth]{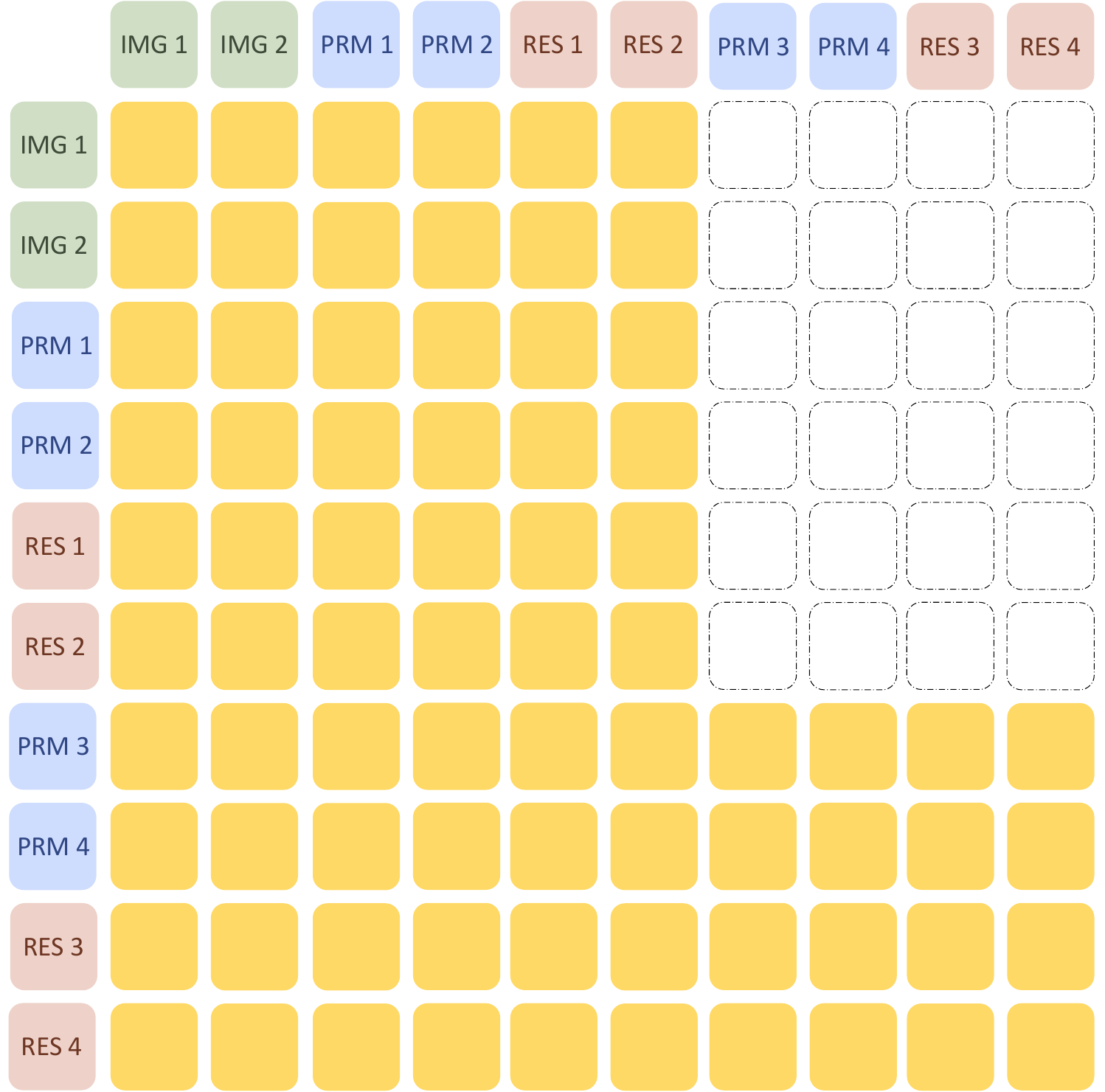}
        \caption{Dialogue Causal Mask}
        \label{fig:dia_ar_attn}
      \end{subfigure}
    \hfill
    \begin{subfigure}[b]{0.325\textwidth}
        \centering
        \includegraphics[width=\textwidth]{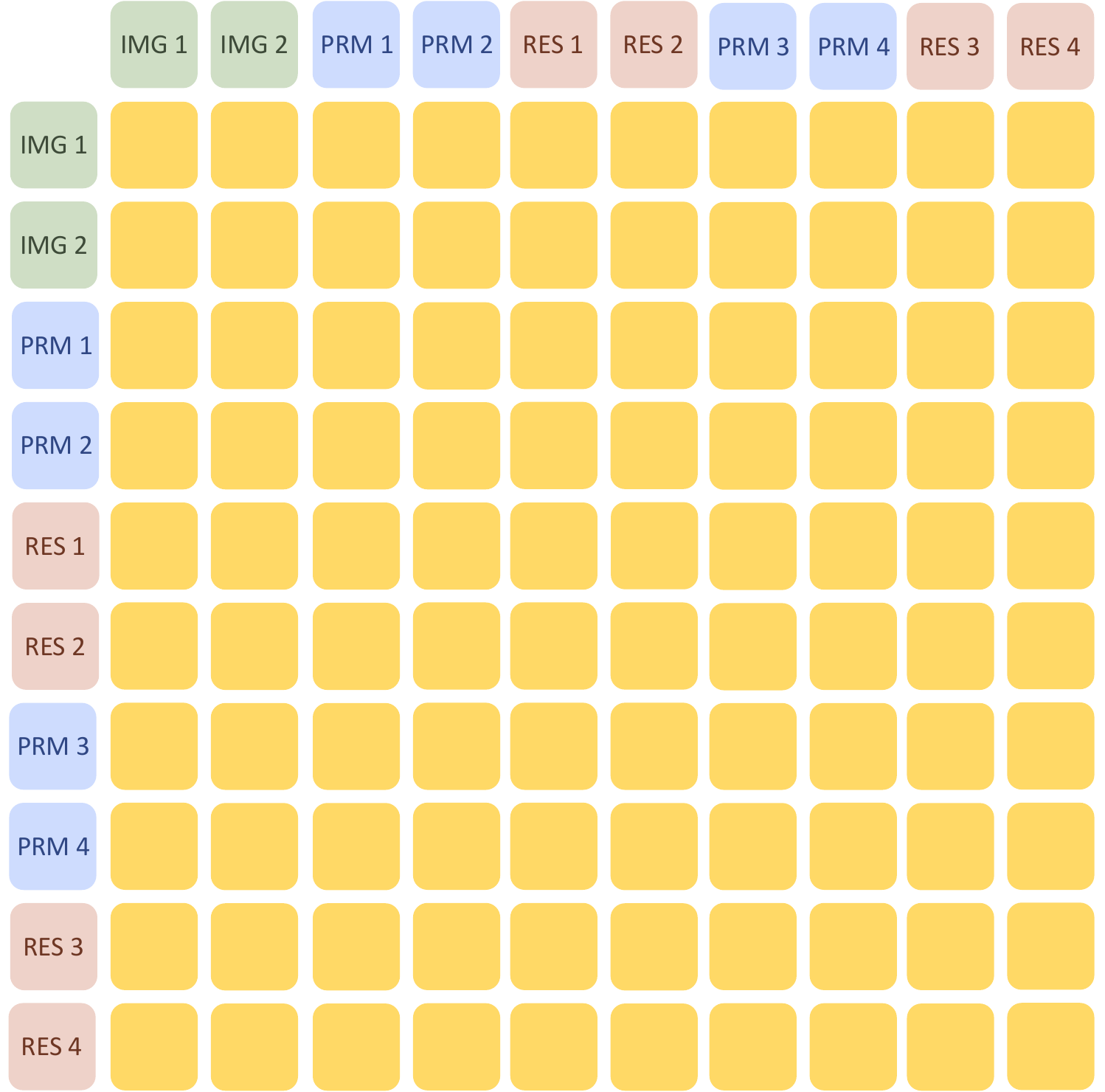}
        \caption{No Mask}
        \label{fig:bidirec_attn}
      \end{subfigure}
    \caption{\textbf{Overview of Attention Masks.} (a) Standard causal mask used in autoregressive models like Qwen2-VL and LLaMA3-V, where tokens attend only to themselves and previous tokens. (b) Dialogue causal mask allowing full attention within turns while maintaining causality between turns. (c) Bidirectional attention in LLaDA-V, enabling tokens to attend to all tokens in the sequence. Note: In the figure, PRM represents prompt and RES represents response. }
  \label{fig:attn_masks}
  \end{figure}

In Fig.~\ref{fig:attn_masks}, we summarize the attention masks discussed in this work. Conventional autoregressive MLLMs utilize a standard causal mask, as shown in Fig.~\ref{fig:ar_attn}, which restricts each token's attention to itself and all previous tokens. LLaDA-V explores two additional alternatives: the \emph{Dialogue Causal Mask}, which allows bidirectional attention within each dialogue turn while preserving causality across turns, which effectively aligns with the structure of multi-turn conversations, and the \emph{No Mask} approach, which enables fully bidirectional attention, allowing all tokens to attend to every other token in the sequence. As discussed in Sec.~\ref{sec:ablation_study}, both attention mask strategies demonstrate strong performance. However, the no mask strategy achieves superior results, outperforming the alternative on 7 out of 12 benchmarks. Consequently, we adopt the no mask strategy as the default in LLaDA-V.

\subsection{Case Studies}

\begin{table*}[t!]
    \centering
    \caption{\textbf{Case Study (1)}}
    \label{tab:visualization_demo0}
    \vspace{1pt} 
    \scalebox{1}{
    \begin{adjustbox}{max width=\textwidth}
    \renewcommand{\arraystretch}{1.2} 
    \begin{tabular}{l p{12.5cm} }
    \toprule
    ~~~~User & Please describe the image in detail \\ 
     & \includegraphics[width=.40\textwidth,valign=t]{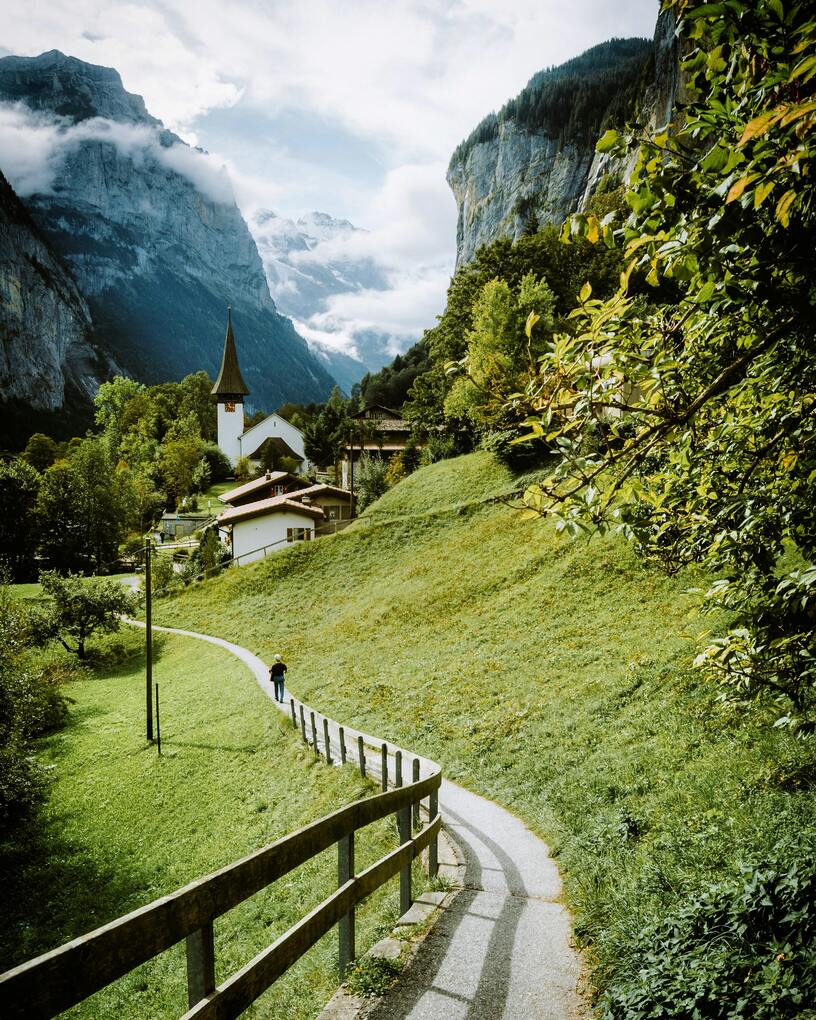}\\ 
    \midrule
    ~~~~LLaDA-V & The image captures a serene and picturesque landscape in the Swiss Alps. The the scene is is a of a lush, green path that meanders through a grassy area. The path is bordered by a wooden fence, and a person can be seen walking along it, adding a sense of scale to the scene.\\ & In the background, a white church with a pointed steeple stands out against the natural backdrop. The church is nestled among a valley, surrounded by green trees and trees. The valley is framed by towering mountains, their peaks shrouded in mist, adding a sense of depth and tranquility to the image.\\ & The sky above is a clear blue, with a few clouds lazily scattered across the sky, contributing to the overall tranquility of the scene. The image is a snapshot of a moment of peace and tranquility, in the natural beauty of the Swiss Alps.\\ & The objects in the image are arranged in layers, with the path in the foreground, the valley in the middle, and the mountains in the background. The path is in the foreground, and the church is in the distance, contributing to the overall tranquility of the scene. The image is a snapshot of a moment of peace and tranquility, in the natural beauty of the Swiss Alps. \\
    \bottomrule
    \end{tabular}
    \end{adjustbox}
    }
\end{table*}

\begin{table*}[t!]
    \centering
    \caption{\textbf{Case Study (2)}}
    \label{tab:visualization_demo1}
    \vspace{1pt} 
    \scalebox{1}{
    \begin{adjustbox}{max width=\textwidth}
    \renewcommand{\arraystretch}{1.2} 
    \begin{tabular}{l p{12.5cm} }
    \toprule
    ~~~~User & How many people are there in the picture? \\ 
     & \includegraphics[width=.40\textwidth,valign=t]{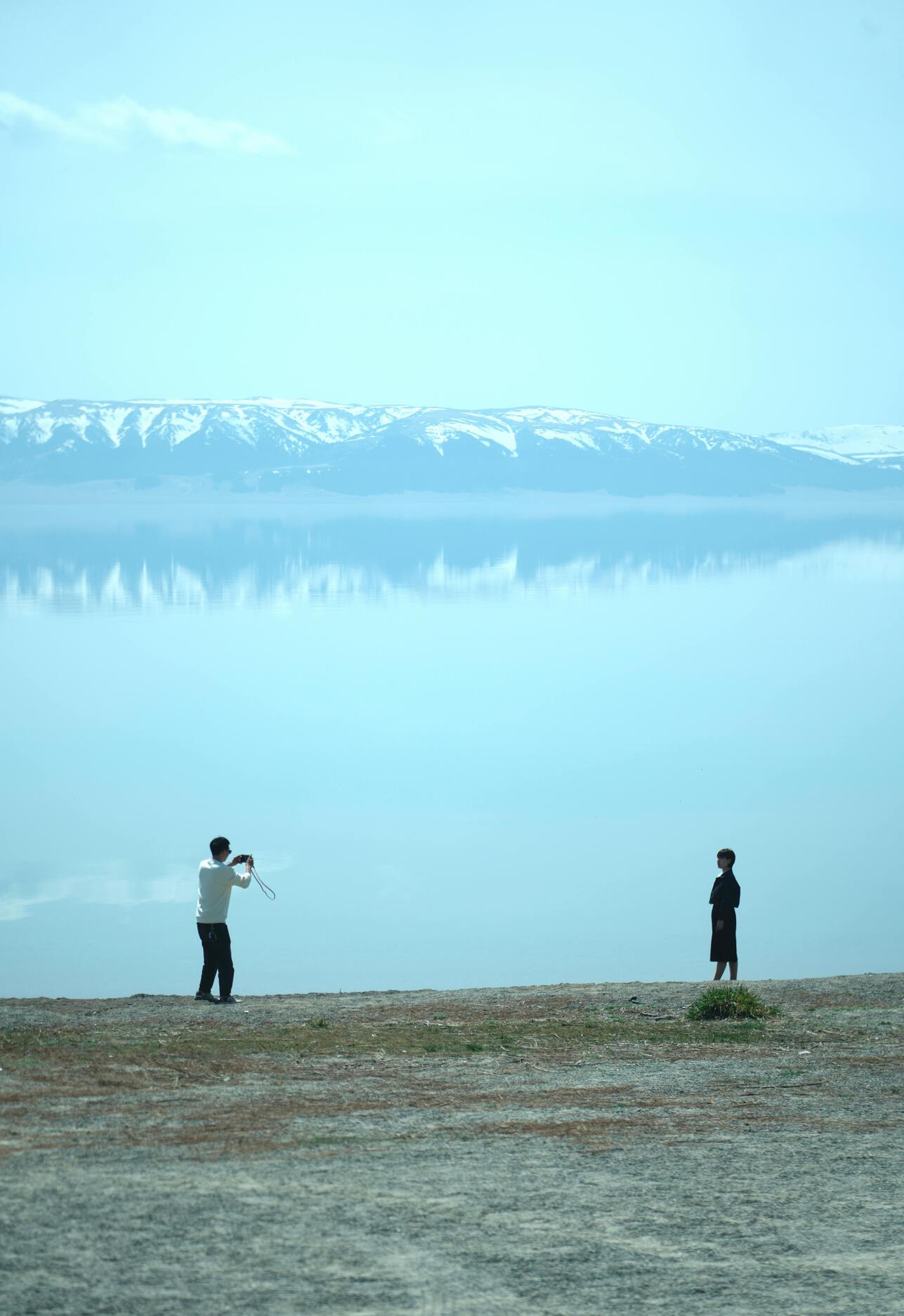}\\ 
    \midrule
    ~~~~LLaDA-V & To determine the number of people in the picture, we need to carefully observe the image and count each individual present.\\ & 1. Observation of the image shows a scenic landscape with a lake of water surrounded by snow-capped mountains. The are two individuals visible in the image. One person is taking a photograph, and the other is standing still. \\ &  2. Counting the individuals:   - The person person taking a photograph is located on the left side of the image.   - The person standing still is located on the right side of the image.\\ & 3. Conclusion:   - By counting these two individuals, we confirm that there are indeed two people present in the picture.\\ & Therefore, the detailed answer is: \\ & The image shows a scenic landscape with a lake of water surrounded by snow-capped mountains. There are two individuals visible in the image.\\ & - We see one person is taking a photograph on the left side of the image. - There is another person standing on the right side of the image.\\ & By counting these two individuals, we confirm that there are 2 people in the picture.\\ & Therefore, the number of people in the picture is 2. \\
    \bottomrule
    \end{tabular}
    \end{adjustbox}
    }
\end{table*}

\begin{table*}[t!]
    \centering
    \caption{\textbf{Case Study (3)}}
    \label{tab:visualization_demo2}
    \vspace{1pt} 
    \scalebox{1}{
    \begin{adjustbox}{max width=\textwidth}
    \renewcommand{\arraystretch}{1.2} 
    \begin{tabular}{l p{12.5cm} }
    \toprule
    ~~~~User & Is this picture real? \\ 
     & \includegraphics[width=.70\textwidth,valign=t]{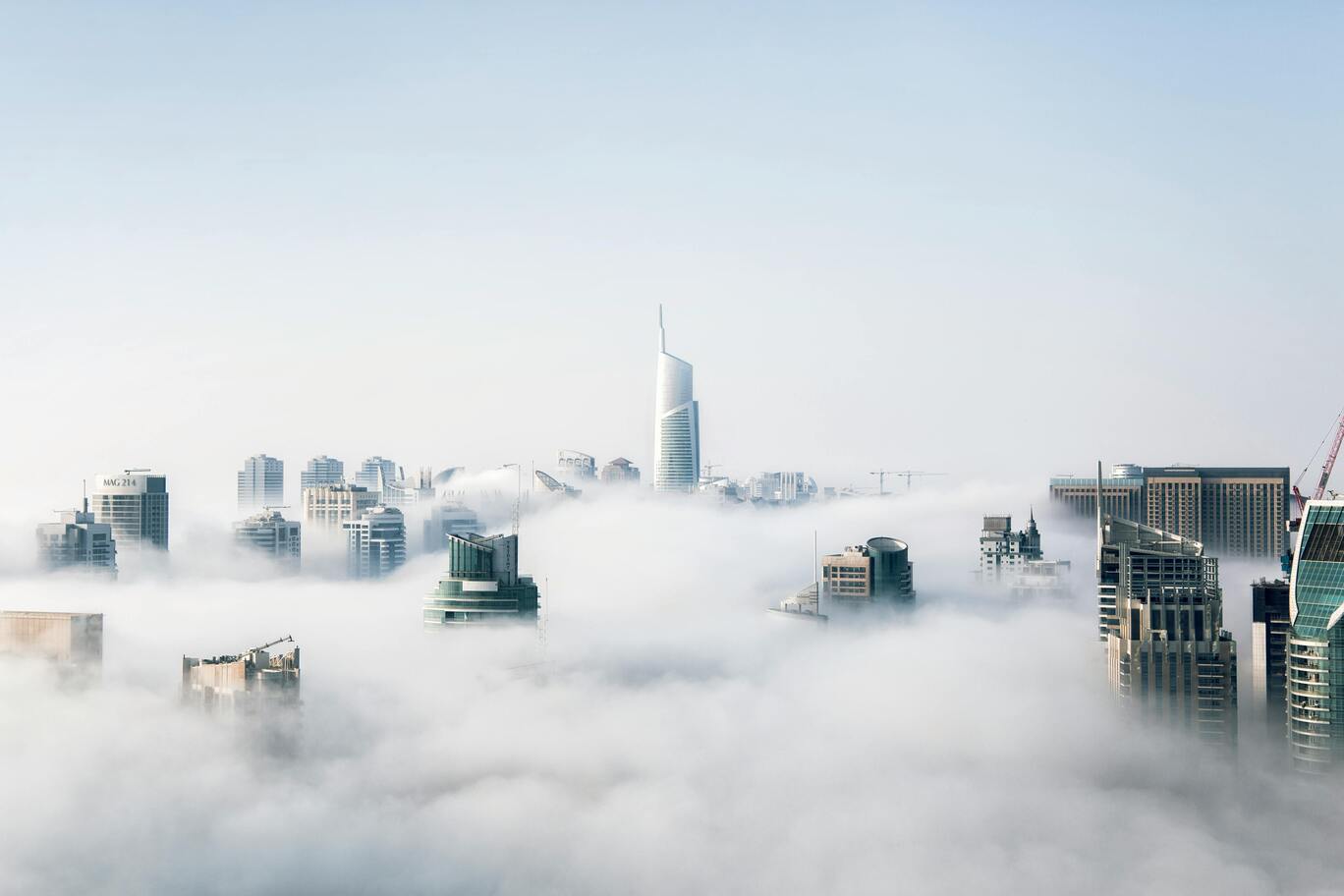}\\ 
     \midrule
    ~~~~LLaDA-V & To determine if the picture is real, we need to analyze several aspects of the image: \\ & 1. \textbf{Resolution and Detail}: The image shows a cityscape with high resolution and detail. The buildings have distinct outlines, and the fog is visible in detail, suggesting a high-quality photograph. \\ & 2. \textbf{Lighting and Shadows}: The lighting in the image appears natural, with shadows cast by the buildings indicating a strong light source, likely the sun. This is consistent with real-world lighting.\\ & 3. \textbf{Color and Contrast}: The colors in the image are vibrant and natural, with the contrast between the buildings and the fog being well-defined. This suggests that the image has been captured with a camera.\\ & 4. \textbf{Composition and Perspective}: The composition of the image is well-balanced, with the cityscape framed in a way that draws the viewer's attention. The perspective is also realistic, consistent with a real photograph.\\ & 5. \textbf{Artistic Elements}: There are no obvious signs of digital manipulation or artifacts, further supporting the conclusion that this is a real photograph.\\ & Given these observations, it is reasonable to conclude that the image is indeed a real photograph.\\ & Therefore, the answer is: Yes, the picture is real.\\
    \bottomrule
    \end{tabular}
    \end{adjustbox}
    }
\end{table*}

\end{document}